\documentclass[a4paper,fleqn]{cas-sc}

\usepackage{booktabs}
\usepackage[authoryear,longnamesfirst]{natbib}
\usepackage{amsmath,amssymb,amsfonts}
\usepackage{graphicx}
\usepackage{textcomp}
\usepackage[dvipsnames]{xcolor}
\usepackage{subcaption}
\usepackage{bm}
\usepackage{mathtools}
\usepackage{multirow}
\usepackage{algorithm,algpseudocode}
\usepackage{varwidth}
\usepackage{siunitx}
\usepackage{enumitem}
\usepackage{hyperref}

\newcommand{\tabitem}{~~\llap{}~~}

\DeclareSIUnit{\sample}{sample}

\ExplSyntaxOn
\NewDocumentCommand{\makervect}{m}
{
	\seq_set_split:Nnn \l_tmpa_seq { , } { #1 }
	\begin{matrix}
		\seq_use:Nn \l_tmpa_seq { & }
	\end{matrix}
}
\ExplSyntaxOff

\ExplSyntaxOff

\def\tsc#1{\csdef{#1}{\textsc{\lowercase{#1}}\xspace}}
\tsc{WGM}
\tsc{QE}

\begin{document}
\let\WriteBookmarks\relax
\def\floatpagepagefraction{1}
\def\textpagefraction{.001}

\shorttitle{A Comparative Study of Rapidly-exploring Random Tree Algorithms Applied to Ship Trajectory Planning and Behavior Generation}

\shortauthors{T. Tengesdal, T. A. Pedersen, T. A. Johansen}

\title [mode = title]{A Comparative Study of Rapidly-exploring Random Tree Algorithms Applied to Ship Trajectory Planning and Behavior Generation}

\tnotemark[1]

\tnotetext[1]{The work was supported by the Research Council of Norway through the Autoship SFI, project number 309230, and by Kongsberg Maritime through the University Technology Center on Ship Performance and Cyber Physical Systems.}



\author[1,2]{Trym Tengesdal}[orcid=0000-0001-8182-3292]
\ead{trym.tengesdal@ntnu.no}
\credit{Conceptualization, Methodology, Software, Validation, Investigation, Writing - Original Draft}
\cormark[1]
\cortext[1]{Corresponding author}

\author[3]{Tom Arne Pedersen}
\ead{tom.arne.pedersen@dnv.com}
\credit{Conceptualization, Writing - Review \& Editing}

\author[1,2]{Tor Arne Johansen}[orcid=0000-0001-9440-5989]
\ead{tor.arne.johansen@ntnu.no}
\credit{Conceptualization, Writing - Review \& Editing, Supervision, Project administration, Funding acquisition}

\affiliation[1]{organization={Autoship Centre for Research-based Innovation (SFI), Norwegian University of Science and Technology (NTNU)},
	addressline={O. S. Bragstads Plass 2D},
	city={Trondheim},
	postcode={7034},
	country={Norway}}

\affiliation[2]{organization={Department of Engineering Cybernetics, NTNU Norwegian University of Science and Technology (NTNU)},
	addressline={O. S. Bragstads Plass 2D},
	city={Trondheim},
	citysep={}, 
	postcode={7034},
	country={Norway}}

 \affiliation[3]{organization={Det Norske Veritas (DNV)},
	addressline={Veritasveien 1},
	city={Høvik},
	citysep={}, 
	postcode={1363},
	country={Norway}}



\begin{abstract}
Rapidly Exploring Random Tree (RRT) algorithms, notably used for nonholonomic vehicle navigation in complex environments, are often not thoroughly evaluated for their specific challenges. This paper presents a first such comparison study of the variants Potential-Quick RRT* (PQ-RRT*), Informed RRT* (IRRT*), RRT*, and RRT, in maritime single-query nonholonomic motion planning. Additionally, the practicalities of using these algorithms in maritime environments are discussed and outlined. We also contend that these algorithms are beneficial not only for trajectory planning in Collision Avoidance Systems (CAS) but also for CAS verification when used as vessel behavior generators.

Optimal RRT variants tend to produce more distance-optimal paths but require more computational time due to complex tree wiring and nearest neighbor searches. Our findings, supported by Welch’s $t$-test at a significance level of $\alpha=0.05$, indicate that PQ-RRT* slightly outperform IRRT* and RRT* in achieving shorter trajectory length but at the expense of higher tuning complexity and longer run-times. Based on the results, we argue that these RRT algorithms are better suited for smaller-scale problems or environments with low obstacle congestion ratio. This is attributed to the curse of dimensionality, and trade-off with available memory and computational resources.
\end{abstract}



\maketitle

\section{Introduction}
\subsection{Background}
Trajectory planning is an important aspect of ship autonomy, not only in Collision Avoidance Systems (CAS) for safe and efficient voyage, but also for the safety assurance of the former. To verify CAS safety and compliance with the International Regulations for Preventing Collision at Sea (COLREG) \citep{IMO1972}, it will be necessary to conduct simulation-based testing in a diverse set of scenarios \citep{Pedersen2020tsbv}. The scenarios must cover varying difficulties with respect to grounding hazards or static obstacles, ships with uncertain kinematics and intentions, and environmental disturbances. Here, it will be important to develop methods for generating interesting and hazardous obstacle vessel behavior scenarios with sufficient variety for the CAS testing. Up until now, this has proved to be a challenging problem not yet solved.

For planning trajectories, roadmap methods such as graph-based search algorithms A* \citep{Hart1968} provide guarantees of finding the optimal solution, if it exists, given its considered grid resolution. However, they do not scale effectively with the state space dimensionality and the problem size, further being unable to consider kinodynamic constraints unless e.g. hybrid A*-variants are used. This often makes it necessary to post-process the solution through smoothing or using optimization-based methods to generate feasible and optimal trajectories. Sampling-based motion planning algorithms such as RRT here provide a good foundation for tackling complex environments with non-convex grounding hazards, without requiring a discretization of the state space, with RRT*-variants \citep{Karaman2010} being \textit{asymptotically optimal}, guaranteeing a probability of finding the optimal solution converging towards unity, if it exists, as the number of iterations approaches infinity. The regular RRT-variants are suboptimal due to the Voronoi bias toward future expansion as opposed to iterative graph rewiring based on node costs in RRT* \citep{Lindemann2004irdi}. However, an underlooked aspect of RRTs is that their inherent randomness makes them promising for trajectory scenario generation, being flexible with respect to biased state space exploration. Different variants can be employed for generating different ship behavior scenarios which can cover their maneuvering space.

\subsection{Previous Work}
Collision-free trajectory planning is a well-studied topic, and we refer the reader to review studies as in \citep{Vagale2021a, Huang2020} for extensive summaries on the various methods proposed previously, and to \citep{Noreen2016} for an extensive review on RRT planning dated 2016. This brief review focuses on core versions of RRT having been proposed previously, in addition to ship scenario generation and falsification, respectively. The RRT algorithms are \textit{single-query}, and thus they are tailored for planning trajectories or paths from a start position to a specified goal. For information on multi-query planning, the reader is referred to e.g. Probabilistic Roadmap methods (PRM) \cite{Kavraki1996probabilistic}.

\subsubsection{Rapidly-exploring Random Trees}
The first baseline RRT planner was introduced by \cite{Lavalle1998}, with the core concept of incrementally sampling configurations or nodes in the obstacle-free space, and wiring of the tree towards these configurations if the trajectory segments in between are collision-free. However, baseline RRT is only \textit{probabilistically complete} in the sense that the probability of the planner finding a solution approaches $1$ as the number of iterations approaches infinity. This issue was addressed in \cite{Karaman2010} with RRT*, giving \textit{asymptotically optimality} guarantees by introducing optimal selection of node parents during tree wiring through nearest neighbor search, and also by rewiring the tree after a new node has been inserted. However, the RRT-based planning still suffered from slow convergence. Since then, a multitude of approaches has been proposed to remedy this \citep{Noreen2016comparison}, e.g. Smart-RRT* in \cite{Nasir2013} and IRRT* \citep{Gammell2014}. The former variant optimizes the found solutions by connecting visible collision-free nodes and utilizing them in the optimized solutions as beacons from which biased samples are drawn at a given percentage. The intelligent sampling procedure, however, only yields improved local convergence to solutions in the vicinity of the current best solution. IRRT* employs a hyperellisoid sampling heuristic from a subset of the planning space, formed after an initial solution is found, and which reduces in volume as the planner improves the current best solution. This variant provides formal linear convergence guarantees, although only given the assumption of no obstacles.

More recently, Potential Quick RRT* (PQ-RRT*) has been proposed \citep{Li2020pq}, as a combination of the Artificial Potential Field (APF) based RRT* in \citep{Qureshi2016potential}, and Quick-RRT* in \citep{Jeong2019quick}. As RRT* is inherently biased towards the exploration of the obstacle-free configuration space, the APF method adds exploitation features through a goal-biased sample adjustment procedure, while the Quick-RRT* gives an improved convergence rate through ancestor consideration in the wiring and re-wiring. As mentioned, there exist a significant number of other variants in the literature, that e.g. add bi-directional tree growth \citep{Klemm2015rrt}, utilize learning-based steering functionality \citep{Chiang2019} or combine A* with RRT* in a learning-based fashion \citep{Wang2020nrrt}. The trade-off between exploitation and exploration in RRT was addressed without nonholonomic system consideration in \citep{Lai2019bgel}. A large literature review on sampling methods utilized in RRTs was given in \citep{Veras2019systematic}, which sheds light on the multitude of variants that have been proposed over the years. In the present work, we only focus on the core versions of directional RRT that have gained traction in the field, namely RRT, RRT*, IRRT*, and PQ-RRT*. 


\subsubsection{Scenario Generation}
This paper also demonstrates a proof-of-concept usage of RRTs for maritime ship behavior scenario generation. We note that this is not new in other domains, as scenario generation and falsification of safety-critical systems have been proposed in e.g. \citep{Dang2008ssse} for testing nonlinear systems subject to disturbances, in \citep{Tuncali2019rrtt} for generating initial car vehicle states that yield boundary cases where the automated vehicle can no longer avoid collision, and in \citep{Koschi2019cesf} for adaptive cruise control falsification to search for hazardous leading vehicle behaviors leading to rear-end collisions.

What has typically been done in previous work within the maritime domain is to consider constant behaviors for vessels involved in a scenario \citep{Minne2017, Pedersen2022, Torben2022, Bolbot2022automatic}. \cite{Torben2022} used a Gaussian Process to estimate how CAS scores concerning safety and the COLREG, which guides the selection of scenarios to test the system at hand, based on its confidence level of having covered the parameter space describing the set of scenarios. \cite{Zhu2022a} proposed an Automatic Identification System (AIS) based scenario generation method. Here, AIS data was analyzed and used to estimate Probability Density Functions (PDFs) describing the parameters of an encounter, such as distances between vessels, their speeds, and bearings. The PDFs were then used to generate a large number of scenarios for testing CAS algorithms. The goal was to increase the test coverage for such systems, over that which is possible with only expert-designed and real AIS data scenarios. Again, generated vessels all follow constant velocity, which does not always reflect true vessel behavior in hazardous encounters. Furthermore, one can not expect all vessels in a given situation to broadcast information using AIS, making AIS-generated scenarios partially incomplete sometimes.

\cite{Porres2020sbts} used the Deep Q Network (DQN) for building a scenario test suite, based on using a neural network to score the performance of randomly generated scenarios. The performance is calculated based on geometric two-ship COLREG compliance and the risk of collision, where the score is used to determine if a given scenario is eligible for simulation and test suite inclusion. The approach should, however, be refined to account for more navigational factors such as grounding hazards in the performance evaluation. Recently, \cite{Bolbot2022automatic} introduced a method for finding a reduced set of relevant traffic scenarios with land and disturbance consideration, through Sobol sequence sampling, filtering of scenarios based on risk metrics, and subsequent similarity clustering. Again, constant behavior is assumed for the vessels, but the process of identifying hazardous scenarios shows promise. 

\subsection{Contributions}
The paper is the first comparison of the previously proposed RRT, RRT*, IRRT*, and Potential Quick-RRT* (PQ-RRT*) together, applied in a complex maritime environment for single-query directional planning with consideration of nonholonomic ship dynamics. This is contrary to the disregard for vehicle dynamics and often simpler and regular environments that have typically been used for testing and comparing RRTs in a lot of previous work.


The planners are compared in Monte Carlo simulations for cases of varying complexity, with consideration of nonholonomic ship dynamics. The trajectory length results produced by PQ-RRT* compared to the other RRT planners are analyzed for statistical significance using Welch`s Student $t$-test. The chosen variants RRT, RRT*, IRRT* and PQ-RRT* are considered as they represent core improvements of the RRT algorithm over the last 25 years, which we deemed the most interesting to compare. Comparisons of fusions of RRT* with path complete methods such as A* as in e.g. \cite{Wang2020nrrt}, bi-directional RRTs, and RRT variants with alternative and more sophisticated steering mechanisms \citep{Chiang2019} are outside the scope of this work. The same goes for the multitude of sampling strategies proposed in the literature \citep{Veras2019systematic}. In addition to the comparison study, the paper also provides guidelines and a discussion around practicalities when applying such algorithms for ship trajectory planning, which can be of use to researchers and practitioners in the field. 

As a side contribution, we argue through proof-of-concept cases that RRT-based planners are beneficial for vessel test scenario generation due to their rapid generation of initially feasible, although not necessarily optimal, trajectories. As we do not necessarily require that obstacle vessels follow optimal trajectories, they provide a viable approach for the fast generation of random vessel scenarios used in CAS benchmarking. RRTs can also be used to generate more realistic ship intention scenarios that can be exploited in intention-aware CAS such as the Probabilistic Scenario-based Model Predictive Control (PSB-MPC) \citep{Tengesdal2024fr, Tengesdal2022_gpu}. Lastly, the RRTs can be used in frameworks as in \citep{Bolbot2022automatic} for finding relevant scenarios, where the RRT sampling heuristics can be tailored to the considered navigational factors. 

\subsection{Outline}
The article is structured as follows. Preliminaries on ship dynamics and control are given in Section \ref{sec:preliminaries}, and background on trajectory planning and RRT variants in Section \ref{sec:rrt}. Notes on practical aspects to consider when applying RRTs for planning are given in Section \ref{sec:practicalities}, whereas results from applying RRTs for ship behavior generation and trajectory planning are given in Section \ref{sec:results}. Lastly, conclusions are summarized in Section \ref{sec:conclusion}.

\section{Preliminaries}\label{sec:preliminaries}
\subsection{Notation}
Vectors and matrices are written as boldface symbols. The Euclidean norm is written as $||\bm{x}||$ for a vector $\bm{x} \in \mathbb{R}^{n}$, with $n$ denoting the vector space dimension. Similarly, a matrix is written as $\bm{M} \in \mathbb{R}^{n \times m}$ and has $n$ rows and $m$ columns.


\subsection{Ship Dynamics}
When planning motion trajectories over longer time horizons, there is seldom a need to consider high-fidelity vehicle models, as modeling errors and disturbances will compound substantially. However, the model should as a minimum take the vehicle's kinodynamical constraints into account. Thus, without loss of generality, we here employ a kinematic ship model on the form $\dot{\bm{x}}(t) = f(\bm{x}(t), \bm{u}(t))$ with $\bm{x} = [x, y, \chi, U]^T \in \mathbb{R}^{n_x}$ consisting of the planar ship position, course, and speed, with input $\bm{u} = [\chi_d, U_d]^T \in \mathbb{R}^{n_u}$ consisting of course and speed autopilot references, and thus $n_x = 4$, $n_u = 2$. The equations of motion are given by
\begin{gather}
\begin{aligned}
    \dot{x} &= U \cos(\chi) \\
    \dot{y} &= U \sin(\chi) \\
    \dot{\chi} &= \frac{1}{T_{\chi}} (\chi_{d} - \chi) \\
    \dot{U} &= \frac{1}{T_{U}} (U_{d} - U) \\
\end{aligned}
\label{eq:prediction_model}
\end{gather}
with time constants $T_{\chi} > 0$ and $T_{U} > 0$ depending on the ship type. The ship maneuverability constraints $U_{min} \leq U \leq U_{max}$ and $|\dot{\chi}| \leq r_{max}$ on the minimum speed $U_{min}$, maximum speed $U_{max}$ and maximum turn rate $r_{max}$ of the ship are considered to ensure kinodynamic feasibility. 

RRTs are capable of considering arbitrarily complex ship dynamics and disturbance models, but we again note that the chosen model yields reduced computational requirements and is suitable for planning trajectories covering larger distances. Over large timespans, external disturbances, and modeling errors will compound substantially over the planning horizon, yielding a low benefit from using high-fidelity vessel models. Note that for lower-level motion control systems used for tracking the RRT trajectory output, it will be important to take the ship dynamics into account. Also, when using RRT to generate random obstacle ship scenarios, one seldom has access to the detailed ship model and motion control system employed by the target ship, further making this model suitable.

\subsection{Line-of-Sight Guidance}\label{subsec:los}
To enable lightweight steering functionality in the RRTs and allowing for orientation or course control, we employ Line-of-Sight (LOS) guidance \citep{Breivik2008} for steering the ship from a waypoint segment from $\{\bm{p}_1, \bm{p}_2\}$, where $\bm{p}_1 = [x^{wp}_1, y^{wp}_1]^T \in \mathbb{R}^2$ is the planar waypoint position. To steer the ship along the straight waypoint segment and towards $\bm{p}_2$, the LOS method first finds the path tangential angle
\begin{equation}
	\theta_p = \text{atan2}(y_{2}^{wp} - y_1^{wp}, x_{2}^{wp} - x_1^{wp})
	\label{eq:path_tangential_angle}
\end{equation}
where $\mathrm{atan2}: \mathbb{R} \rightarrow (-\pi, \pi]$ is the four-quadrant arctangent function. Then, the path deviation $\bm{\epsilon}(t)$, referenced to the path-fixed frame, is computed as
\begin{equation}
	\bm{\epsilon} = \bm{R}(\theta_p)^T(\bm{p}(t) - \bm{p}_1)
	\label{eq:path_dev}
\end{equation}
with the rotation matrix $\bm{R}(\theta_p)$ given by
\begin{equation}
	\bm{R}(\theta_p) = \begin{bmatrix}
		\cos(\theta_p) & -\sin(\theta_p) \\
		\sin(\theta_p) & \cos(\theta_p)
	\end{bmatrix}
	\label{eq:rot_alpha_k}
\end{equation}
The path deviation $\bm{\epsilon}(t) = [s(t), e(t)]^T$ consists of the along-track error $s(t)$ and cross-track error $e(t)$, respectively, where the latter is used with the path tangential angle $\theta_p$ to set the desired course-over-ground (COG) $\chi_{d}(t)$ as 
\begin{equation}
    \chi_{d}(t) = \theta_p + \arctan\left(-\frac{e(t)}{\Delta}\right)
    \label{eq:desired_COG}
\end{equation}
where $\Delta$ is the look-ahead distance that determines how fast the ship will turn towards the straight line segment. The desired speed-over-ground (SOG) $U_{d}(t)$ can, in general, be varying, but is typically set to a constant in the case of nominal trajectory planning and generation. Without loss of generality, we do not consider environmental disturbances. However, note that compensations for slowly varying disturbances can be considered by adding an integral term in \eqref{eq:desired_COG}. See \citep{Breivik2008} for illustrations and more information. 

\section{Trajectory Planning Using Rapidly-exploring Random Trees}\label{sec:rrt}
\subsection{Problem Definition}
This section formally defines the considered motion planning problem and the RRT-based algorithms compared in this work: Standard RRT \citep{Lavalle1998}, RRT* \citep{Karaman2010}, IRRT* \citep{Gammell2014} and Potential Quick RRT* (PQ-RRT*) \citep{Li2020pq}. As mentioned in the introduction, bi-directional variants such as RRT*-connect \citep{Klemm2015rrt} and combinations of RRT with e.g. path-complete methods such as A* are outside the scope of this work. The following text does not provide an in-depth introduction to each algorithm, and the reader is thus referred to the cited references for more details.

We consider the own-ship motion model $\Dot{\bm{x}} = f(\bm{x}(t), \bm{u}(t))$ as given by \eqref{eq:prediction_model} with state $\bm{x}(t) \in \mathcal{\bm{X}}$ and input $\bm{u}(t) \in \mathcal{\bm{U}}$, belonging to the configuration space $\mathcal{\bm{X}} \subset \mathbb{R}^{n_x}$ and input space $\mathcal{\bm{U}} \subset \mathbb{R}^{n_u}$, respectively. Let $\mathcal{\bm{X}}_{obst}$ denote the forbidden static obstacle space and $\mathcal{\bm{X}}_{free} = \mathcal{\bm{X}} \backslash \mathcal{\bm{X}}_{obst}$ the static obstacle-free configuration space. 

The optimal trajectory planning problem is stated as follows. Let $\bm{z}_{start} \in \mathcal{\bm{X}}_{free}$ and $\bm{z}_{goal} \in \mathcal{\bm{X}}_{free}$ be the initial and final own-ship states, respectively. Further let $\bm{\sigma}_d: [0, T_{plan}] \rightarrow \mathcal{\bm{X}}$ be a non-trivial trajectory from start to goal. Then, the objective of the RRT-based planning algorithms is to find the optimal and feasible desired trajectory $\bm{\sigma}_d^{*}$ defined through
\begin{equation}
    \bm{\sigma}_d^{*} = \textrm{arg} \hspace{0.1em} \underset{\bm{\sigma}_d}{\textrm{min}} \hspace{0.3em} \left( c(\bm{\sigma}_d) \hspace{0.1em} | \hspace{0.1em} \bm{\sigma}_d(0) = \bm{x}_{start}, \hspace{0.3em} \bm{\sigma}_d(T_{plan}) = \bm{x}_{end}, \hspace{0.5em} \forall t \in \{0, T_{plan}\}, \hspace{0.3em} \bm{\sigma}_d(t) \in \mathcal{X}_{free} \right)
\end{equation}
where $T_{plan}$ is the trajectory duration and $c(\cdot): \mathcal{\bm{X}} \rightarrow \mathbb{R}$ the planning problem cost function. In summary, the required inputs and resulting output of the RRT planners are 
\begin{itemize}[leftmargin=50pt, rightmargin=0pt]
    \item [\textbf{Inputs}] $\bm{z}_{start}$, $\bm{z}_{goal}$ and Electronic Navigational Chart (ENC): Starting state, goal state, and ENC containing grounding hazard information, respectively.
    \item [\textbf{Output}] Collision-free trajectory plan $\bm{\sigma}_d$ starting in $\bm{z}_{start}$ and ending in $\bm{z}_{goal}$.
\end{itemize}
The output trajectory from the RRT-based planners can be generated offline or online depending on the application and is typically provided as a reference trajectory for the lower-level motion control system onboard the ship to track. 

\subsection{Core RRT-functionality}
To solve the described trajectory planning problem, RRT-based planners incrementally and randomly grow a tree $\mathcal{T} = (\mathcal{V}, \mathcal{E})$ consisting of a state vertex or node set $\mathcal{V} \subset \mathcal{\bm{X}}_{free}$ that are connected through the directed edges in $\mathcal{E} \subseteq \mathcal{V} \times \mathcal{V}$. Asymptotic optimality in the number of iterations for the planned trajectory is guaranteed in RRT* and its variants through the parent-cost dependent wiring and rewiring of the tree in order to find new minimal cost parents \citep{Karaman2011}. The standard RRT method does, however, not have this property. We note that the convergence rate is highly dependent on the node sampling procedures, in addition to the tree re-wiring mechanism. Common functions used in all the RRT-based planners are given below.


1) $\mathrm{Sample}(i)$: Samples a random state $z_{rand} \in \mathcal{\bm{X}}_{free}$. See Section \ref{subsec:sampling} for implementation aspects. When applied to random scenario generation, the sampling can be biased towards scenario-related metrics. 

2) $\mathrm{Dist}(\bm{z}_1, \bm{z}_2)$: Computes the Euclidian distance between two states $\bm{z}_1, \bm{z}_2 \in \mathcal{\bm{X}}$.

3) $\mathrm{Cost}(\bm{z})$: Computes the cost $c(\bm{z}) = c(\bm{z}_{parent}) + \mathrm{Dist}(\bm{z}, \bm{z}_{parent})$ of a new node $\bm{z}$ as the cost of its parent $\bm{z}_{parent}$, plus the Euclidian distance $\mathrm{Dist}(\bm{z}, \bm{z}_{parent})$.

4) $\mathrm{Nearest}(\mathcal{T}, \bm{z})$: Given a tree $\mathcal{T}$ and a state $\bm{z} \in \mathcal{\bm{X}}$, this function finds the nearest node $\bm{z}_{nearest} = \mathrm{Nearest}(\mathcal{T}, \bm{z})$ in the tree by utilizing the $\mathrm{Dist}(\cdot)$ function.


5) $\mathrm{Insert}(\bm{z}_{current}, \bm{z}_{new}, \mathcal{T})$: Given the current node $\bm{z}_{current} \in \mathcal{\bm{X}}$, the function inserts a new node $\bm{z}_{new}$ into the current tree $\mathcal{T}$ by adding it to $\mathcal{V}$, and connects it to $\bm{z}_{current}$ by adding the edge between them to $\mathcal{E}$. 

6) $\mathrm{IsCollisionFree}(\bm{\sigma})$: Checks whether a trajectory $\bm{\sigma}: [0, T_{plan}] \rightarrow \mathcal{\bm{X}}$ is outside the obstacle space $\mathcal{\bm{X}}_{obst}$ for all $t \in \{0, T_{plan}\}$. 

7) $\mathrm{Steer}(z_1, z_2)$: Computes a control input sequence that steers the own-ship from state $\bm{x}(0) = \bm{z}_1$ to $\bm{x}(T) = \bm{z}_2$ using the LOS method outlined in Section \ref{subsec:los}. The result is the final endpoint state $\bm{z}_{new}$ and corresponding trajectory $\bm{\sigma}_{new}: [0, T] \rightarrow \mathcal{X}$ with steering horizon $T \in [T_{min},T_{max}]$. The minimum and maximum steering time  $T_{min}$ and $T_{max}$ are parameters to be adjusted based on the geography. Narrow channels and inland waterways might require lower steering times to avoid collisions, and vice versa for more open sea areas. 

8) $\mathrm{ExtractBestSolution}(\mathcal{T})$: Given the tree $\mathcal{T}$, the function extracts the solution 
\begin{equation}
\bm{\sigma}_d = \textrm{arg} \hspace{0.1em} \underset{\bm{\sigma}}{\textrm{min}} \hspace{0.3em} \left(c(\bm{\sigma}) \hspace{0.1em} | \hspace{0.1em} \bm{\sigma} \in \mathcal{T} \hspace{0.1em} \land \hspace{0.1em} c(\bm{\sigma}) < \infty \right), 
\end{equation}
if any. Otherwise, the algorithm will report failure. The solution trajectory is valid if the corresponding leaf node is inside an acceptance radius $R_{a}$ of the goal $\bm{z}_{goal}$, and thus its cost is finite. 

The $\mathrm{Steer}$ function enables the planner to consider the own-ship dynamics by using the motion model $f(\bm{x}(t), \bm{u}(t))$ for simulating a trajectory from a state $z_1$ to $z_2$. The speed and turn rate (course rate) are saturated to within the considered vessel minimum and maximum speeds $\{U_{min}, U_{max}\}$ and maximum turn rate $r_{max}$ for this particular model, respectively.

To incorporate dynamic obstacle collision avoidance, one can employ a joint simulator as in \cite{Chiang2018} in the steering together with adding virtual obstacles for striving towards COLREG compliance, or utilize biased sampling methods as in e.g. \citep{Enevoldsen2021colregs}. However, this will not be considered in the present work. 

For reductions in computational effort, we add the functionality that the planner attempts direct goal growth every $\Delta_{goal}$ iterations through the following function:

9) $\mathrm{DirectGoalGrowth}(\mathcal{T}, \bm{z}_{goal})$: Finds the nearest neighbor $\bm{z}_{nearest} = \mathrm{Nearest}(\mathcal{T}, \bm{z}_{goal})$ of the goal state, and attempts to steer the ship towards the state using $\mathrm{Steer}(\bm{z}_{nearest}, \bm{z}_{goal})$ with a sufficiently large steering time, commonly a multiple of $T_{max}$. This returns a new node $\bm{z}_{new}$, which is added to the tree $\mathcal{T}$ if the steering is successful.

The baseline RRT algorithm can be described by these functions and is outlined in Algorithm \ref{alg:rrt}. Here, $N_{iter}^{max}$ is the maximum number of allowable iterations and $\Delta_{goal}$ the iterations between each attempt of $\mathrm{DirectGoalGrowth}$. In addition to the maximum iteration constraint, we also under the hood put a constraint on the maximum number of nodes $N_{node}^{max}$ allowable in the RRT planners.
\begin{algorithm}
    \caption{RRT}\label{alg:rrt}
    \begin{algorithmic}[1]
        \Require \{Initial state $\bm{z}_{init}$, goal state $\bm{z}_{goal}$ and ENC \}
        \State $\mathcal{T} \leftarrow (\emptyset, \emptyset)$
        \State $\mathcal{T} \leftarrow \mathrm{Insert}(\emptyset, \bm{z}_{init}, \mathcal{T})$
        \For{$i = 1 : N_{iter}^{max}$}
            \If{$ i \quad \% \quad \Delta_{goal} == 0$}
                $\mathrm{DirectGoalGrowth}(\mathcal{T}, \bm{z}_{goal})$
            \EndIf
            \State $\bm{z}_{rand} \leftarrow \mathrm{Sample}(i)$
            \State $\bm{z}_{nearest} \leftarrow \mathrm{Nearest}(\mathcal{T}, \bm{z}_{rand})$
            \State $(\bm{z}_{new}, \bm{\sigma}_{new}) \leftarrow \mathrm{Steer}(\bm{z}_{nearest}, \bm{z}_{rand})$
            \If{$\mathrm{IsCollisionFree}(\bm{\sigma}_{new})$}
                \State $\mathcal{T} \leftarrow \mathrm{Insert}(\mathcal{T}, \bm{z}_{min}, \bm{z}_{new})$
            \EndIf 
        \EndFor
        \State $\bm{\sigma}_d \leftarrow \mathrm{ExtractBestSolution}(\mathcal{T})$
    \end{algorithmic}
\end{algorithm}

\subsection{RRT*}
Specific to the RRT* variant are the $\mathrm{NearestNeighbors}$, $\mathrm{FindParent}$ and $\mathrm{Rewire}$ functions, which give the algorithm probabilistic asymptotic convergence properties. The RRT* algorithm can be described fully by the functions 1-12 and is outlined in Algorithm \ref{alg:rrt_star}. 

10) $\mathrm{NearestNeighbors}$: The function $\mathrm{NearestNeighbors}(\mathcal{T}, \bm{z}, n) = \mathcal{V} \cap \mathrm{Reachables}(\bm{z}, l)
$ finds the nearest neighbours $\mathcal{Z}_{near}$ in $\mathcal{V}$ of the state $\bm{z} \in \mathcal{\bm{X}}$, by calculation of the set $\mathrm{Reachables}(\bm{z}, l) = \{\bm{z}^{\prime} \in \mathcal{\bm{X}} : \mathrm{Dist}(\bm{z}, \bm{z}^{\prime}) \leq l \land \mathrm{Dist}(\bm{z}^{\prime}, \bm{z}) \leq l\}$. The distance threshold is given as $l = l(n)$ and found such that the $\mathrm{Reachables}(\bm{z}, l)$ contains a ball with volume $\gamma \mathrm{log}(n)/n$ for an appropriate parameter $\gamma$ \citep{Karaman2010}. To reduce tree size and computation time, we enforce that the distance to the nearest node must satisfy $\mathrm{Dist}(\bm{z}_{new}, \bm{z}_{nearest}) \geq d_{node}^{min}$, and consider a maximum number of neighbors of $N_{nn}^{max} = 10$ for all RRT* variants.

11) $\mathrm{FindParent}(\mathcal{Z}_{near}, \bm{z}_{nearest}, \bm{z}_{new})$: Finds the parent of the currently considered node $\bm{z}_{new} \in \mathcal{\bm{X}}$ that gives the minimal $\mathrm{Cost}(\bm{z}_{new})$ out of the nearest node set $\mathcal{Z}_{near}$ and the nearest node by distance $\bm{z}_{nearest}$ \citep{Karaman2010}. 

12) $\mathrm{Rewire}(\mathcal{T}, \mathcal{Z}_{near}, \bm{z}_{min}, \bm{z}_{new})$: Checks if the tree $\mathcal{T}$ can be rewired to give a lower cost for $\bm{z}_{new}$ than the current parent $\bm{z}_{min}$ by connecting it to a minimal cost node in $\mathcal{Z}_{near}$. See \citep{Karaman2010} for more information. 

Nearest neighbor extraction is a crucial part of RRT*, and is highly dependent on the search parameter $\gamma$, for which \citep{Noreen2016comparison} provide some tuning considerations. We note that this parameter must be scaled based on the problem size. If dimensions $> 2$ are considered, with state entries of different units and scales, it will be important to normalize the states. For Euclidean distance-based search in two or three dimensions, this is, however, a necessity.
\begin{algorithm}
    \caption{RRT*}\label{alg:rrt_star}
    \begin{algorithmic}[1]
        \Require \{Initial state $\bm{z}_{init}$, goal state $\bm{z}_{goal}$ and ENC\}
        \State $\mathcal{T} \leftarrow (\emptyset, \emptyset)$
        \State $\mathcal{T} \leftarrow \mathrm{Insert}(\emptyset, \bm{z}_{init}, \mathcal{T})$
        \For{$i = 1 : N_{iter}^{max}$}
            \If{$ i \quad \% \quad \Delta_{goal} == 0$}
                $\mathrm{DirectGoalGrowth}(\mathcal{T}, \bm{z}_{goal})$
            \EndIf
            \State $\bm{z}_{rand} \leftarrow \mathrm{Sample}(i)$
            \State $\bm{z}_{nearest} \leftarrow \mathrm{Nearest}(\mathcal{T}, \bm{z}_{rand})$
            \State $(\bm{z}_{new}, \bm{\sigma}_{new}) \leftarrow \mathrm{Steer}(\bm{z}_{nearest}, \bm{z}_{rand})$
            \If{$\mathrm{IsCollisionFree}(\bm{\sigma}_{new})$}
                \State $\mathcal{Z}_{near} \leftarrow \mathrm{NearestNeighbors}(\mathcal{T}, \bm{z}_{new})$
                \State $\bm{z}_{min} \leftarrow \mathrm{FindParent}(\mathcal{Z}_{near}, \bm{z}_{nearest}, \bm{z}_{new})$
                \State $\mathcal{T} \leftarrow \mathrm{Insert}(\mathcal{T}, \bm{z}_{min}, \bm{z}_{new})$
                \State $\mathcal{T} \leftarrow \mathrm{Rewire}(\mathcal{T}, \mathcal{Z}_{near}, \bm{z}_{min}, \bm{z}_{new})$
            \EndIf 
        \EndFor
        \State $\bm{\sigma}_d \leftarrow \mathrm{ExtractBestSolution}(\mathcal{T})$
    \end{algorithmic}
\end{algorithm}

\subsection{IRRT*}
IRRT* varies from the baseline RRT* only in the fact that, once an initial solution is found with cost $c_{best}$, an admissible informed sampling heuristic is used onwards. The heuristic is formed from an elliptical domain given by the $\bm{x}_{start}$ and $\bm{x}_{goal}$ as focal points, the theoretical minimum cost $c_{min} = ||\bm{p}_{goal} - \bm{p}_{start}||_2$ and current best solution cost $c_{best}$, with $\bm{p}_{goal}$ and $\bm{p}_{start}$ being the planar position parts $\bm{x}_{start}$ and $\bm{x}_{goal}$. It is shown that this heuristic enables \textit{focused} planning towards $\bm{x}_{goal}$, as opposed to the Voronoi bias property inherited in RRT and RRT* that lead to planning towards all points in the state space. The informed sample is drawn from a hyperellipsoid as follows:
\begin{equation}
    \bm{z}_{ell} = \bm{C}\bm{L}\bm{z}_{ball} + \bm{z}_{centre},
    \label{eq:informed_sample}
\end{equation}
with $\bm{z}_{centre} = (\bm{x}_{start} + \bm{x}_{goal})/2$ and $\bm{L} \in \mathbb{R}^{n_x \times n_x}$ being the Cholesky decomposition of the matrix $\bm{S} = \bm{L}\bm{L}^T \in \mathbb{R}^{n_x \times n_x}$, which is given by 
\begin{equation}
    \bm{S} = \mathrm{diag}\Bigl\{\frac{c_{best}^2}{4}, \frac{c_{best}^2 - c_{min}^2}{4}, ...,  \frac{c_{best}^2 - c_{min}^2}{4} \Bigr\},
\end{equation}
and which defines the ellipsoid
\begin{equation}
    (\bm{z} - \bm{z}_{centre})^T \bm{S} (\bm{z} - \bm{z}_{centre}) \leq 1
\end{equation}
The rotation matrix $\bm{C}$ from the hyperellipsoidal frame to the world frame can be found by solving the Wahba problem \citep{Gammell2014}. However, when sampling planar positions, it can be directly found as a 2D rotation matrix with angle $\theta_g = \mathrm{atan2}(y_{goal} - y_{start}, x_{goal} - x_{start})$. The sampling is in this case reduced to 
\begin{equation}
    \begin{aligned}
        \bm{C}_{2D} &= \begin{bmatrix}
            cos(\theta_g) & -sin(\theta_g) \\
            sin(\theta_g) & cos(\theta_g)
        \end{bmatrix} \\
        \bm{p}_{ball} &\sim \mathrm{Uniform}(\bm{p}; \bm{0}_{2\times 1}, \bm{1}_{2\times 1}) \\
        \bm{p}_{ell} &= \bm{C}_{2D}\bm{L}_{2D}\bm{p}_{ball} + \bm{p}_{centre}\\
        \bm{z}_{ell} &= [\bm{p}_{ell}^T, 0, 0]^T,
    \end{aligned}
    \label{eq:planar_informed_sample}
\end{equation}
where $\mathrm{Uniform}(x; a, b)$ is a uniform distribution with interval limits $a$ and $b$.
The informed sampling procedure is described below, with the IRRT*-variant being described in Algorithm \ref{alg:informed_rrt_star}.

13) $\mathrm{InformedSample}(i)$: Given the current best cost $c_{best}$, sample a new state using \eqref{eq:planar_informed_sample}. If $c_{best} = \infty$, use the baseline $\mathrm{Sample}(i)$.

\begin{algorithm}
    \caption{IRRT*}\label{alg:informed_rrt_star}
    \begin{algorithmic}[1]
        \Require \{Initial state $\bm{z}_{init}$, goal state $\bm{z}_{goal}$ and ENC\}
        \State $\mathcal{T} \leftarrow (\emptyset, \emptyset)$
        \State $\mathcal{T} \leftarrow \mathrm{Insert}(\emptyset, \bm{z}_{init}, \mathcal{T})$
        \For{$i = 1 : N_{iter}^{max}$}
            \If{$ i \quad \% \quad \Delta_{goal} == 0$}
                $\mathrm{DirectGoalGrowth}(\mathcal{T}, \bm{z}_{goal})$
            \EndIf
            \State $\bm{z}_{rand} \leftarrow \mathrm{InformedSample}(i)$
            \State $\bm{z}_{nearest} \leftarrow \mathrm{Nearest}(\mathcal{T}, \bm{z}_{rand})$
            \State $(\bm{z}_{new}, \bm{\sigma}_{new}) \leftarrow \mathrm{Steer}(\bm{z}_{nearest}, \bm{z}_{rand})$
            \If{$\mathrm{IsCollisionFree}(\bm{\sigma}_{new})$}
                \State $\mathcal{Z}_{near} \leftarrow \mathrm{NearestNeighbors}(\mathcal{T}, \bm{z}_{new})$
                \State $\bm{z}_{min} \leftarrow \mathrm{FindParent}(\mathcal{Z}_{near}, \bm{z}_{nearest}, \bm{z}_{new})$
                \State $\mathcal{T} \leftarrow \mathrm{Insert}(\mathcal{T}, \bm{z}_{min}, \bm{z}_{new})$
                \State $\mathcal{T} \leftarrow \mathrm{Rewire}(\mathcal{T}, \mathcal{Z}_{near}, \bm{z}_{min}, \bm{z}_{new})$
            \EndIf 
        \EndFor
        \State $\bm{\sigma}_d \leftarrow \mathrm{ExtractBestSolution}(\mathcal{T})$
    \end{algorithmic}
\end{algorithm}

\subsection{PQ-RRT*}
The Potential-Quick RRT* combines the features of P-RRT* and Q-RRT* \citep{Li2020pq}, which involves sample adjustments using a goal-based potential field attractive force, and ancestor consideration in the tree growth for path length reduction, respectively. The potential-field based sample adjustment procedure is shown in Algorithm \ref{alg:adjust_sample}, whereas the $\mathrm{Ancestry}$ functionality are described by the following procedures:

14) $\mathrm{Ancestor}(\mathcal{T}, \bm{z}, \phi)$: Given the tree $\mathcal{T}$, a node $\bm{z}$ and depth parameter $\phi > 0$, the $\phi$-th parent of $\bm{z}$ is returned. If the tree is not deep enough, no ancestor is returned. 

15) $\mathrm{Ancestry}(\mathcal{T}, \mathcal{Z})$: Given the tree $\mathcal{T}$ and node set $\mathcal{Z}$, it returns $\mathcal{Z}_{ancestors} = \bigcup\limits_{\bm{z}\in \mathcal{Z}} \biggl[\bigcup\limits_{l=1}^d \mathrm{Ancestor}(\mathcal{T}, \bm{z}, l)\biggr]$, and $\emptyset$ if the tree depth is 0.

16) $\mathrm{PQRewire}(\mathcal{T}, \mathcal{Z}_{near}, \bm{z}_{min}, \bm{z}_{new})$: Same as $\mathrm{Rewire}$, except from the fact that the ancestry of $\bm{z}_{new}$ is also considered as a rewiring node.

The PQ-RRT* is described in Algorithm \ref{alg:pqrrt}. In this work, we consider a depth-level of $\phi = N_{ancestry}$ in the $\mathrm{Ancestry}$ procedure, whereas a constant depth level of $1$ is used in the $\mathrm{PQRewire}$ method, to reduce computational effort. Note that the algorithm as proposed in \cite{Li2020pq} was not tested with kinodynamical constraints and motion planning in the steering, which will induce substantially higher run-times in the algorithm due to the ancestor consideration in the $\mathrm{FindParent}$ and $\mathrm{PQ-Rewire}$ routines. On the other hand, better convergence properties are expected due to the PQ features. The sample adjustment procedure uses a hazard clearance parameter $d_{margin}$ that must be set based on acceptable margins, the ship type, and possibly other factors. We note that the selection of the PQ-RRT* specific parameters is non-trivial, and no guidelines for their selection were provided in \cite{Li2020pq}. The authors highlighted this as a potential limitation of the method.
\begin{algorithm}
    \caption{$\mathrm{AdjustSample}$}\label{alg:adjust_sample}
    \begin{algorithmic}[1]
        \Require \{Sampled state $\bm{z}_{rand}$, goal state $\bm{z}_{goal}$ \}
        \State $\bm{z}_{prand} \leftarrow \bm{z}_{rand}$
        \For{$k = 1 : N_{sa}^{max}$}
            \State $\bm{F}_{att} \leftarrow (\bm{z}_{goal} - \bm{z}_{prand})$
            \State $d_{min} \leftarrow \mathrm{DistanceNearestObstacle}(\mathcal{X}_{obst}, \bm{z}_{prand})$
            \If{$d_{min} < d_{margin}$}
                \State $\mathbf{return} \quad  \bm{z}_{prand}$
            \Else 
                \State $\bm{z}_{prand} \leftarrow \bm{z}_{prand} + \lambda \frac{\bm{F}_{att}}{||\bm{F}_{att}||}$
            \EndIf
        \EndFor
        \State $\mathbf{return} \quad  \bm{z}_{prand}$
    \end{algorithmic}
\end{algorithm}

\begin{algorithm}
    \caption{PQ-RRT*}\label{alg:pqrrt}
    \begin{algorithmic}[1]
        \Require \{Initial state $\bm{z}_{init}$, goal state $\bm{z}_{goal}$ and ENC\}
        \State $\mathcal{T} \leftarrow (\emptyset, \emptyset)$
        \State $\mathcal{T} \leftarrow \mathrm{Insert}(\emptyset, \bm{z}_{init}, \mathcal{T})$
        \For{$i = 1 : N_{iter}^{max}$}
            \If{$ i \quad \% \quad \Delta_{goal} == 0$}
                $\mathrm{DirectGoalGrowth}(\mathcal{T}, \bm{z}_{goal})$
            \EndIf
            \State $\bm{z}_{rand} \leftarrow \mathrm{Sample}(i)$
            \State $\bm{z}_{rand} \leftarrow \mathrm{AdjustSample}(\bm{z}_{rand}, \bm{z}_{goal})$
            \State $\bm{z}_{nearest} \leftarrow \mathrm{Nearest}(\mathcal{T}, \bm{z}_{rand})$
            \State $(\bm{z}_{new}, \bm{\sigma}_{new}) \leftarrow \mathrm{Steer}(\bm{z}_{nearest}, \bm{z}_{rand})$
            \If{$\mathrm{IsCollisionFree}(\bm{\sigma}_{new})$}
                \State $\mathcal{Z}_{near} \leftarrow \mathrm{NearestNeighbors}(\mathcal{T}, \bm{z}_{new})$
                \State $\mathcal{Z}_{ancestors} \leftarrow \mathrm{Ancestry}(\mathcal{T}, \mathcal{Z}_{near})$
                \State $\mathcal{Z}_{union} \leftarrow \mathcal{Z}_{near} \cup \mathcal{Z}_{ancestors}$
                \State $\bm{z}_{min} \leftarrow \mathrm{FindParent}(\mathcal{Z}_{union}, \bm{z}_{nearest}, \bm{z}_{new})$
                \State $\mathcal{T} \leftarrow \mathrm{Insert}(\mathcal{T}, \bm{z}_{min}, \bm{z}_{new})$
                \State $\mathcal{T} \leftarrow \mathrm{PQRewire}(\mathcal{T}, \mathcal{Z}_{near}, \bm{z}_{min}, \bm{z}_{new})$
            \EndIf 
        \EndFor
        \State $\bm{\sigma}_d \leftarrow \mathrm{ExtractBestSolution}(\mathcal{T})$
    \end{algorithmic}
\end{algorithm}

\subsection{Algorithm Pros and Cons}
As it can be non-obvious for the reader to see the immediate main differences between the considered RRT-based planners, we provide Table \ref{tab:pros_cons_rrt} below to summarize the pros and cons of each algorithm. The jagging and meandering tendency of RRT, RRT*, and IRRT* owes to the fact that only a single ancestor is considered in the rewiring, which is also highly dependent on the choice of nearest neighbor search parameter $\gamma$.
\begin{table*}[htp]
	\centering
	\caption{Summary of the considered RRT-based planners.}
	\bgroup
	\def\arraystretch{1.4}%
	\begin{tabular}{c|c|c|c}
		\text{Algorithm} &  Optimal & Pros  & Cons \\
            \hline
		RRT	& No & Can find initial solutions fast & Highly jagged output trajectories \\
            \tabitem  & & Little to no tuning required & Non-focused planning\\
            \hline
		RRT*	& Yes &  Simple to tune & Jagged output trajectories \\
            \tabitem &  & & Non-focused planning \\ 
		\hline
		IRRT*	& Yes & Focused planning & High sample rejection rate in vanilla version \\
            \tabitem & & Simple to tune & Jagged output trajectories \\
            \hline
		PQ-RRT*	& Yes & Focused planning & Non-trivial tuning \\
            \tabitem & & Low jagging and meandering tendency & Higher computational effort required
	\end{tabular}
	\egroup
	\label{tab:pros_cons_rrt}
\end{table*}

\section{Practical Aspects in RRT-based Planning}\label{sec:practicalities}
\subsection{Sampling}\label{subsec:sampling}
\begin{figure}[htp]
    \centering
    \includegraphics[width=0.7\columnwidth,trim={2em, 0em, 2em, 0em},clip]{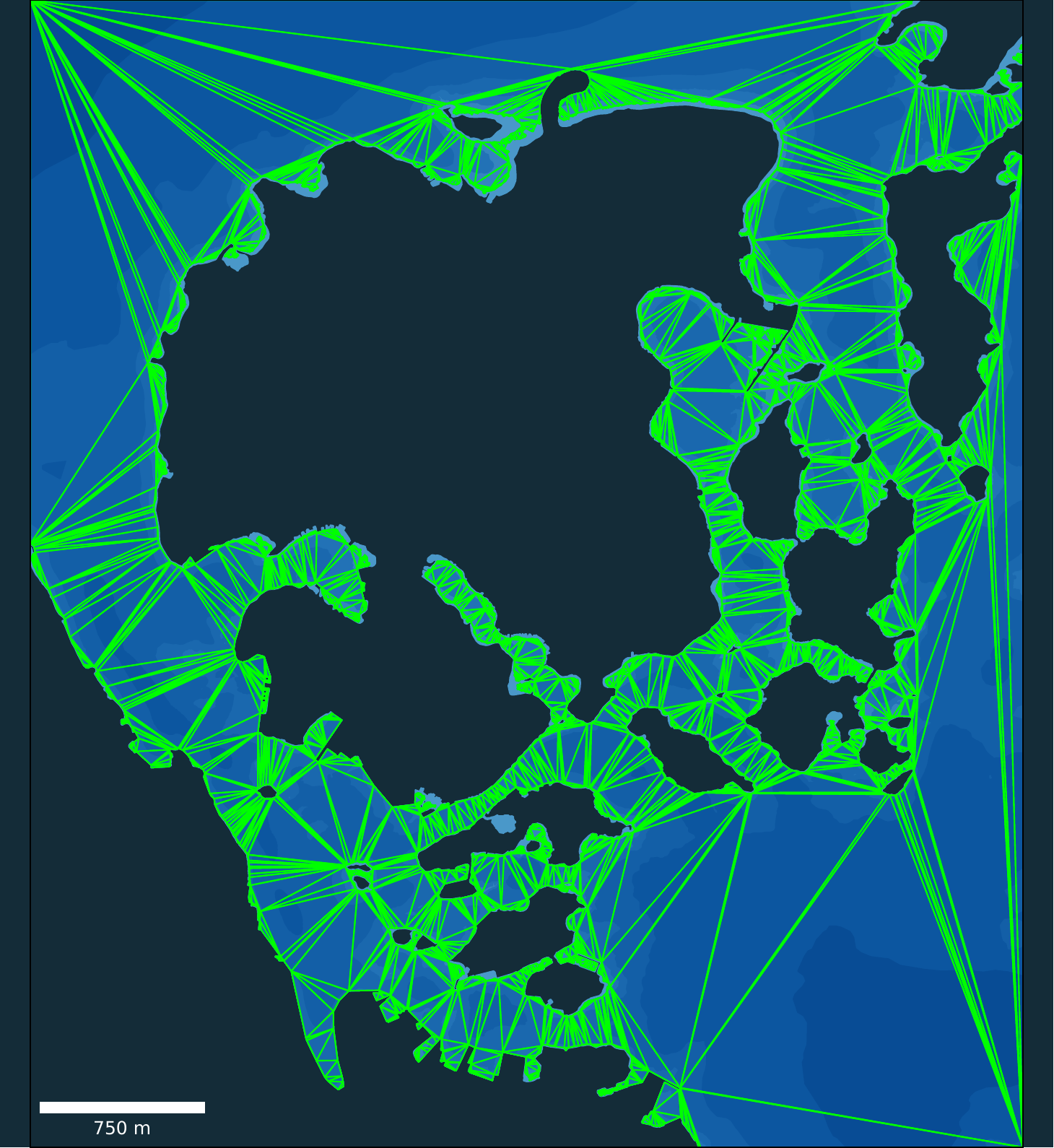}
    \caption{Constrained Delaunay Triangulation for a considered planning area near Stavanger, Norway. Triangles are shown with \textcolor{Green}{green} edges, Land in \textcolor{Blue}{dark blue}, whereas the shore and seabed are shown in variations of \textcolor{RoyalBlue}{brighter blue}. }
    \label{fig:cdt_ex}
\end{figure}

One of the most important parts of RRT is the method by which new states are sampled. Brute force sampling of states $\bm{z}_{rand} \in \mathcal{X}_{free}$ is not recommended, as the rejection rate will be proportional to the volume of $\mathcal{X}_{obst}$ relative to the total space $\mathcal{X}$. Instead, one can for instance create and sample from a Constrained Delaunay Triangulation (CDT), made from the safe sea area that the ship is to voyage within, similar to \citep{Enevoldsen2022sampling}. This gives a set $\Theta_{tri}$ of triangles with indices $j = 1, 2, ..., n_{tri}$, which in this application reduces the sampling of a new state position $\bm{p}_{rand} \in \mathbb{R}^2$ to 
\begin{equation}
    \begin{aligned}
        j^{\prime} & \sim \mathrm{WeightedUniform}(j; 1, n_{tri}) \\
        r_1 & \sim \mathrm{Uniform}(r; 0, 1) \qquad r_2 \sim \mathrm{Uniform}(r; 0, 1) \\
        \bm{p}_{rand} &= (1 - \sqrt{r_1})\bm{A}_{j^{\prime}} + \sqrt{r_1}(1 - r_2) \bm{B}_{j^{\prime}} + \sqrt{r_1}r_2 \bm{C}_{j^{\prime}}
    \end{aligned}
    \label{eq:tri_sample}
\end{equation}
where $(\bm{A}_{j^{\prime}}, \bm{B}_{j^{\prime}}, \bm{C}_{j^{\prime}})$ are the vertices of triangle $j^{\prime}$, and $\mathrm{WeightedUniform}$ is a uniform distribution weighted by the area of each triangle. The random state is then found as $\bm{z}_{rand} = [\bm{p}_{rand}^T, 0, 0]^T$. An illustration of a CDT of the safe sea area used for sampling in the second planning example is shown in Fig. \ref{fig:cdt_ex}. 

When using the sample heuristic in IRRT* bonafide, the same problem of high rejection rates can occur. Here, one can again use CDT, and prune triangles outside the ellipsoidal sampling domain after a new solution is found and the heuristic is put to use. However, this will not be done here.

\subsection{Data Structures}
Another foundation of the RRT algorithms is nearest neighbor searches, used when wiring and re-wiring the tree for RRT*-variants. For spatial queries as is considered here, it is recommended to use R*-tree or R-tree-based data structures \citep{Guttman1984r, Beckmann1990r} that are commonly used for storing spatial objects in e.g. databases. They have $O(log(n))$ complexity for insertion and distance-related queries for $n$ inserted elements. The trees are created such that leaf nodes of the tree hold spatial data, and parents or branching nodes correspond to the minimum bounding box that contains all of its children. With this structure, the R-tree utilizes data-based partitions into boxes of decreasing size as the tree grows. K-d trees have the same complexity as R-trees, and can also be used \citep{Bentley1975multidimensional}. However, note that the bonafide version is suitable only when the nodes are points and does not work well in high dimensions. Once polygons and other objects are introduced for representing the own-ship, static or dynamic hazards, standard k-d trees are also not compatible. The advantages of using R-trees are the efficient memory layout, tree update procedures and spatial nearest neighbor searching. For cases where the tree will not end up with a high number of objects or when few modifications and removals will be made on the go, it can on the other hand be sufficient to use k-d trees \citep{Bentley1975multidimensional}.

These points also apply to the collision checking part, discussed below. Note that it might be worthwhile to consider a Mahalanobis distance metric when it is desired to evaluate the distance with respect to both position and other state variables such as orientation. In this regard, it will be wise to normalize the data before considering distance calculations.

\subsection{Collision Checking}
To accept a new node in the tree $\mathcal{T}$ maintained by the RRT algorithms, collision checks must be performed for each new trajectory segment $\bm{\sigma}_{new}$ resulting from the steering function in several parts of the RRT-algorithms. To ensure feasibility with respect to run-time, it is important to pre-process grounding hazard data using line simplification algorithms such as Ramer-Douglas-Peucker \citep{Ramer1972} to reduce the map data accuracy to the required level. Especially for the maritime domain, one should merge multi polygons arising from hazardous \textit{land}, \textit{shore} and \textit{seabed} objects extracted from Electronic Navigation Charts (ENC) \citep{Blindheim2022_seacharts}, given the considered vessel and its draft. We also recommend removing interior holes of land multi polygons arising from e.g. lakes, as these will naturally not be considered for sea voyages. 

Again, it will also be important in this procedure to use efficient data structures such as R-trees for enabling fast distance computations in collision checking. Adherence to specified hazard clearance margins $d_{safe}$ and control requirements can easily be done by buffering the hazard polygons before running the algorithm. Note that verification of sampled poses along a trajectory being collision-free is not exact, and operation in more confined space with smaller distance margins might require more accurate methods as in e.g. \citep{Zhang2022gccd}.


Note that collision checking requires significantly less effort when only two-point planar waypoint segments are considered, as opposed to a full trajectory segment with kinodynamical constraint consideration. This can be an option when the goal is to rapidly generate waypoints for a ship to follow, as opposed to a full trajectory to track. Some regard to maximum curvature based on the ship turn rate and speed should then be taken into account when accepting waypoint segments. 

\subsection{Steering}
For trajectory planning RRTs, the steering functionality for connecting state pairs $(\bm{z}_1, \bm{z}_2)$ is a crucial part of the tree wiring, especially for non-holonomic systems. Previous methods have proposed e.g. the simplistic Dubin´s path for optimal steering \citep{Karaman2010}, Bezier spline-based connectivity \citep{Yang2014spline}, learning-based steering and collision checking \citep{Chiang2019}, LQR-control-based extension \citep{Perez2012lqr} and lazy steering with a Neural Network (NN) for determining collision-free and steerable node extensions \citep{Yavari2019srrt}. In this work, we apply LOS guidance for steering the ship towards new eligible nodes, as a low-cost solution for creating feasible trajectory segments in the maritime domain. One point that requires care in this context, is the early termination of the LOS guidance if the waypoint segment from $\bm{z}_1$ to a new node $\bm{z}_2$ is passed by, such that the computational cost of numerical integration of the ship dynamics is kept to a minimum.  Further note that the integration time step should be increased in tact with the problem size, also related to the computational cost aspect.

\section{Results}\label{sec:results}
We present three cases in which the considered RRT-variants are compared. The first case demonstrates the usage of RRTs for rapid ship behavior generation, where the goal is to generate kinodynamically feasible trajectories for a specified ship scenario. The second case compares common RRT variants for a smaller planning scenario with a typical local minimum problem. Finally, the third case compares the RRT planners in a larger planning scenario. The simulation framework in \citep{Tengesdal2023_sim} is used as a platform for developing and testing the planners, which allows for the utilization of Electronic Navigational Charts. The algorithms are implemented in the \textit{Rust programming language}, and the executable is run on a MacBook Pro with an Apple Silicon M1 chip. We evaluate the obtained solutions with respect to run-time and trajectory lengths, over $N_{MC} = 100$ Monte Carlo (MC) simulations for each case, where the RRTs are provided with different pseudorandom seeds before each run. Aside from summarizing the planner results, we report the statistical significance of the trajectory length results produced by PQ-RRT* being more distance optimal than other RRT variants or not through Welch`s unequal variance Student`s $t$-test \citep{Ruxton2006unequal}.

A kinematic model with parameters $T_{\chi} = \SI{6}{s}$, $T_U = \SI{6}{s}$, $r_{max} = \SI{10}{\degree\per\second}$, $U_{min} = \SI{0}{knots}$ and $U_{max} = \SI{20}{knots}$ is considered for a ship with length about $\SI{15}{m}$, with LOS guidance parameters $\Delta = \SI{30.0}{m}$. 

The grounding hazards in the environment are buffered with a horizontal clearance parameter $d_{safe}$ of $\SI{0}{m}$ in the first two cases, and  $\SI{5}{m}$ in the last case. This will in general be dependent on the map accuracy, ship type, and application. We consider a time step $\delta_{sim}$ of $\SI{0.5}{s}$ in the first two cases, and $\SI{1.0}{s}$ in the last case, for the RRT kinematic ship model. For the RRTs, we use R-trees to perform nearest neighbor searching and spatial queries, where grounding hazards (land, shore, and relevant seabed) are extracted and merged considering a vessel draft of $\SI{1.0}{m}$. The merged hazards are then used to form a safe sea CDT, from which weighted samples are drawn. 

Key parameters for the RRT-variants used in the first two cases are given in Table \ref{tab:parameters_case_12}. For the first case, we did not find a configuration of PQ-RRT* parameters for the sample adjustment procedure that gave a better result than just disabling the APF-based part as a whole and thus used $N_{sa}^{max} = 0$. All methods sample from the safe sea CDT using \eqref{eq:tri_sample}, except the IRRT* which uses \eqref{eq:informed_sample} after a solution has been found. The PQ-RRT* will adjust the CDT sample using its goal potential field. We note that tuning of the RRT algorithms is non-trivial, and will be specific to the scale considered for the trajectory planning. This is a trade-off between planner run-time, memory requirements, and solution quality. The number of iterations should be adequately high to increase the likelihood of the RRT finding a solution.
\begin{table}[!htp]
	\centering
	\caption{Algorithm parameters for the smaller planning cases.}
	\bgroup
	\def\arraystretch{1.4}%
	\begin{tabular}{c|c|c|c|c}
		\text{Parameter} & \text{PQ-RRT*} & \text{IRRT*} & \text{RRT*} & \text{RRT} \\
		\hline
		$N_{node}^{max}$	& $10000$ & $10000$ & $10000$ & $10000$ \\
            \hline 
            $N_{iter}^{max}$	& $25000$ & $25000$ & $25000$ & $25000$ \\
            \hline 
            $\Delta_{goal}$	& $500$ & $500$ & $500$ & $500$ \\
             \hline 
            $d_{node}^{min}$	& $\SI{5.0}{m}$ & $\SI{5.0}{m}$ & $\SI{5.0}{m}$ & - \\
            \hline 
            $\gamma$ & $\SI{2000.0}{m}$ & $\SI{2000.0}{m}$ & $\SI{2000.0}{m}$ & - \\
            \hline 
            $T_{min} $ & $\SI{1.0}{s}$ & $\SI{1.0}{s}$ & $\SI{1.0}{s}$ & $\SI{1.0}{s}$ \\
		\hline
            $T_{max} $ & $\SI{30.0}{s}$ & $\SI{30.0}{s}$ & $\SI{30.0}{s}$ & $\SI{30.0}{s}$ \\
            \hline
            $R_{a}$	& $\SI{10.0}{m}$ & $\SI{10.0}{m}$ & $\SI{10.0}{m}$ & $\SI{10.0}{m}$ \\
            \hline 
            $\delta_{sim} $ & $\SI{0.5}{s}$ & $\SI{0.5}{s}$ & $\SI{0.5}{s}$ & $\SI{0.5}{s}$ \\
            \hline
            $d_{margin}$ & $\SI{0.1}{m}$ & - & - & - \\ 
            \hline
            $N_{ancestry}$ & $1$ & - & - & - \\ 
            \hline
            $N_{sa}^{max}$ & $0$ & - & - & - \\
            \hline
            $\lambda_{sample}$ & $\SI{1.0}{m}$ & - & - & - \\
	\end{tabular}
	\egroup
	\label{tab:parameters_case_12}
\end{table}

\subsection{Random Vessel Trajectory Generation}\label{subsec:case1}
The first case is a random vessel trajectory generation scenario where the goal is to generate multiple random trajectories from a given start position. After being built, the resulting RRT variant can be queried efficiently through R-tree spatial nearest neighbor search \citep{Guttman1984r}, and used to rapidly sample random ship trajectories for intention-aware CAS or simulation-based testing of CAS. This is useful in the online setting for CAS, but also for interaction data generation to be used by learning-based CAS algorithms. Training of Reinforcement Learning (RL) agents with the \href{https://github.com/Farama-Foundation/Gymnasium}{Gymnasium} framework typically requires a reset of the environment after each terminated or truncated episode. In this context, RRT algorithms can be used for rapidly generating target ship trajectory scenarios after a reset. As an example, PQ-RRT* will, in general, produce more path-optimal solutions than RRT and RRT*, and can thus be used for spawning target ship behavior scenarios with minimal maneuvering, whereas the RRT* or RRT variants can be used for spawning gradually unpredictable maneuvering target ship behaviors that can be considered as outliers. Thus, employing multiple RRT algorithm variants for ship scenario generation can improve the ability of learning-based CAS to generalize, and also enlarge the test coverage in the context of simulation-based CAS verification. 

Note that, in the scenario-generation context the RRT cost function can be designed to e.g. minimize time to collision or converge towards near misses between the random vessel and the own-ship that runs the CAS to be tested \citep{Tuncali2019rrtt}. Furthermore, in the maritime context, COLREG can be misinterpreted and lead to ambiguous and therefore often dangerous situations \citep{Chauvin2008}. Thus, RRTs could be guided towards edge case situations in COLREG where the applicable situation rule(s) are easy to misinterpret. These considerations are a topic for future work. 

\begin{figure}[htp]
    \centering
    \subcaptionbox{PQ-RRT*. \label{fig:pqrrt_scengen}}
    {\includegraphics[width=0.45\columnwidth,trim={0em, 0em, 10em, 0em},clip]{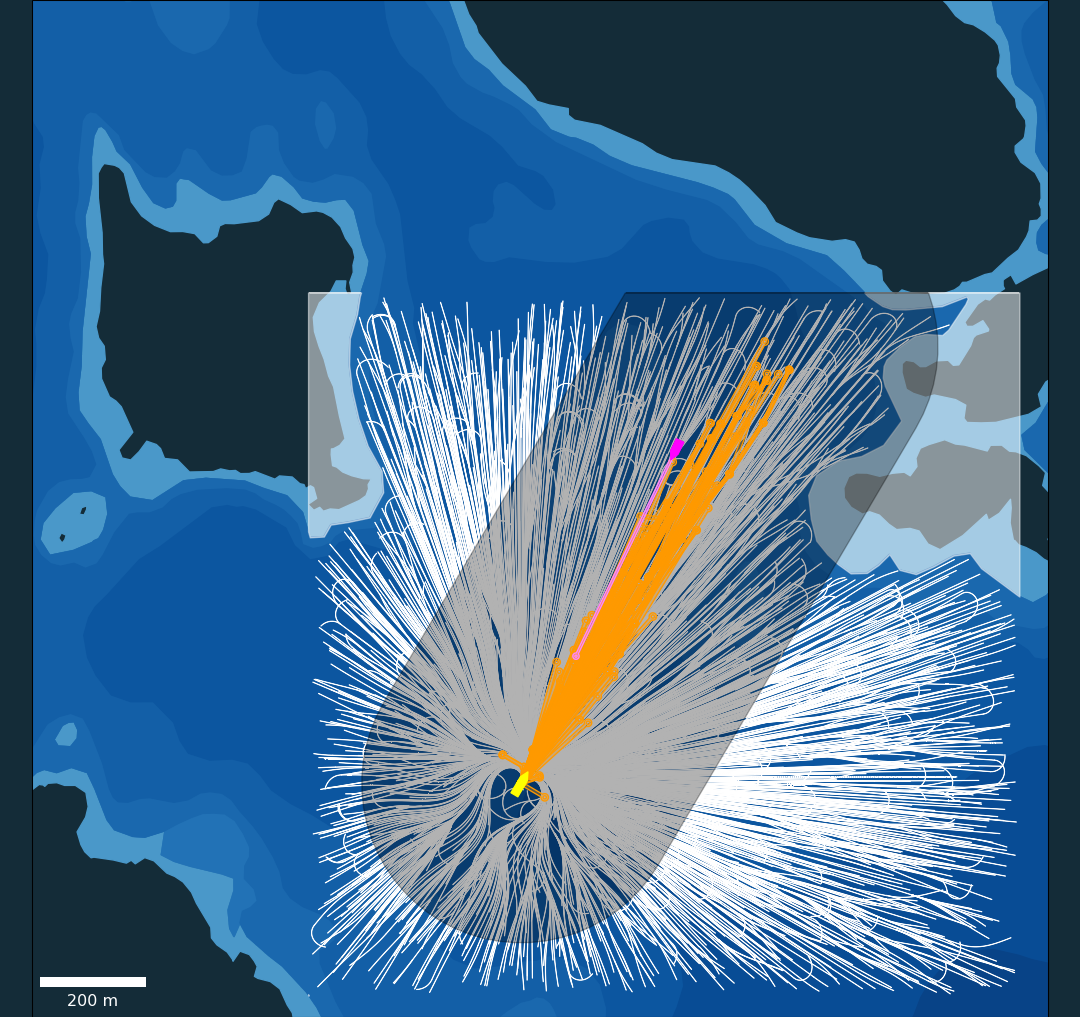}}
    \subcaptionbox{RRT*. \label{fig:rrt_star_scengen}}
    {\includegraphics[width=0.45\columnwidth,trim={0em, 0em, 10em, 0em},clip]{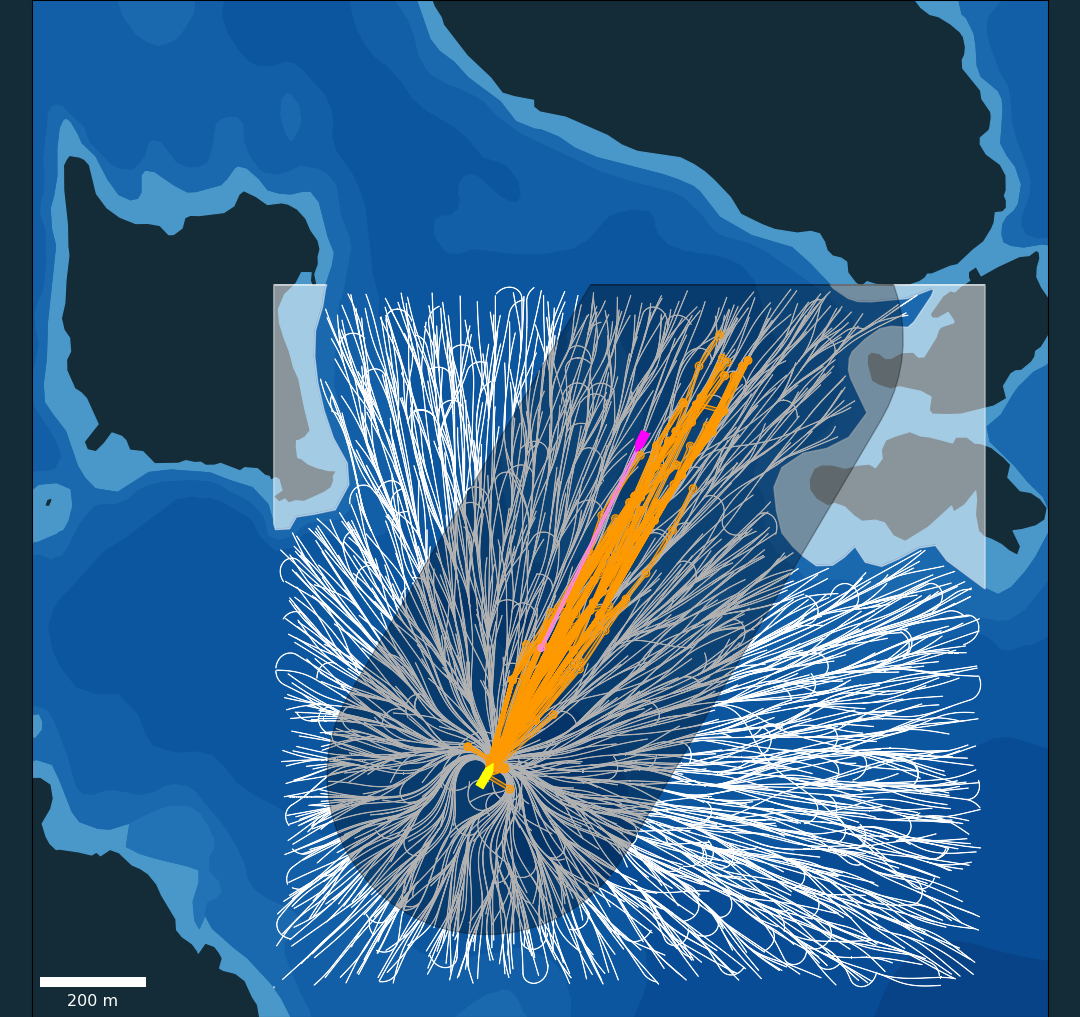}}
    \subcaptionbox{RRT. \label{fig:rrt_scengen}}
    {\includegraphics[width=0.45\columnwidth,trim={0em, 0em, 10em, 0em},clip]{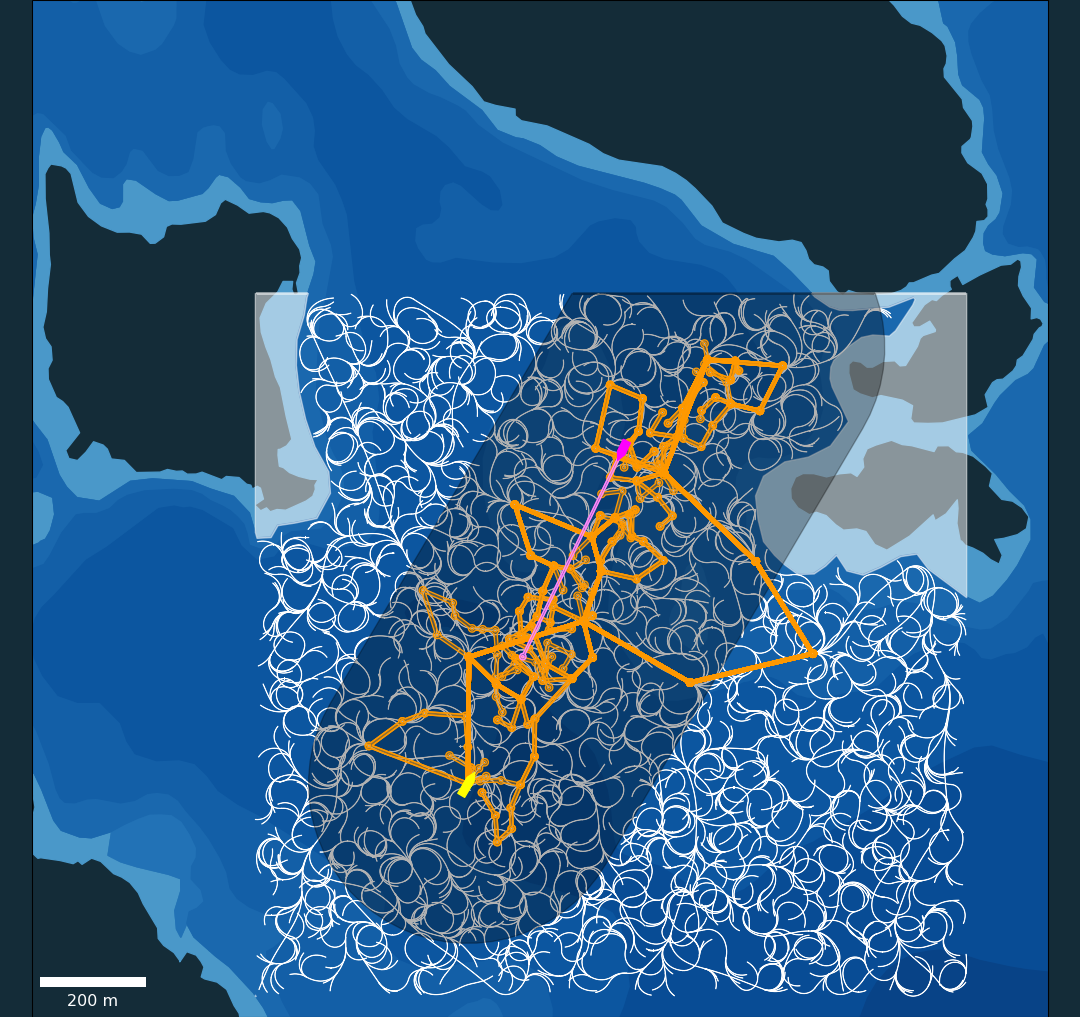}}\\
    \caption{Results from random obstacle ship behavior generation for a CAS head-on situation near Halsn\o ya in Boknafjorden, Norway. The initial obstacle ship position is shown in yellow, the initial own-ship position in \textcolor{CarnationPink}{pink} and its waypoints in \textcolor{Purple}{purple}, the resulting tree in white, sampled obstacle ship behavior waypoint sequences in \textcolor{Orange}{orange} annuluses. The safety buffered hazard boundary is shown in \textcolor{Red}{red}, with the planning bounding box shown in transparent white. }
    \label{fig:scen_gen_ex1}
\end{figure}

Fig. \ref{fig:scen_gen_ex1} shows an example of obstacle ship behavior generation for a CAS head-on situation. To make the figures less dense, we have reduced the maximum allowable tree nodes to $N_{node}^{max}= 4000$ for all planners. In this particular case, we find built RRT, RRT*, and PQ-RRT* behaviors close to a randomly sampled position within a corridor given by the initial own-ship position and course and its maximum travel length over the simulation timespan, i.e.  
\begin{equation}
    \begin{aligned}
        x_{corr} & \sim \mathrm{Uniform}(x; 0, U_{d}^{os}T_{sim}) \\
        y_{corr} & \sim \mathrm{Uniform}(y; -d_{corr}/2, d_{corr}/2) \\
        \theta_{corr} &= \textrm{atan2}(y_{end}^{os} - y_{start}^{os}, x_{end}^{os} - x_{start}^{os}) \\
        \bm{p}_{rand} &= \bm{p}^{os}_{start} + \bm{R}(\theta_{corr}) \begin{bmatrix}
            x_{corr} \\ y_{corr}
        \end{bmatrix},
    \end{aligned}
    \label{eq:corr_sample}
\end{equation}
with $d_{corr}$ as the corridor width parameter, $T_{sim}$ as the total simulation time-span, $U_{d}^{os}$ is the own-ship speed reference, and $\bm{p}_{start}^{os} = [x_{start}^{os}, y_{start}^{os}]^T$ and $(x_{end}^{os}, y_{end}^{os})$ are the own-ship start and end coordinates, respectively. 

\begin{figure}[htp]
    \centering
    \subcaptionbox{PQ-RRT*. \label{fig:pqrrt_scengen2}}
    {\includegraphics[width=0.45\columnwidth,trim={0em, 0em, 0em, 0em},clip]{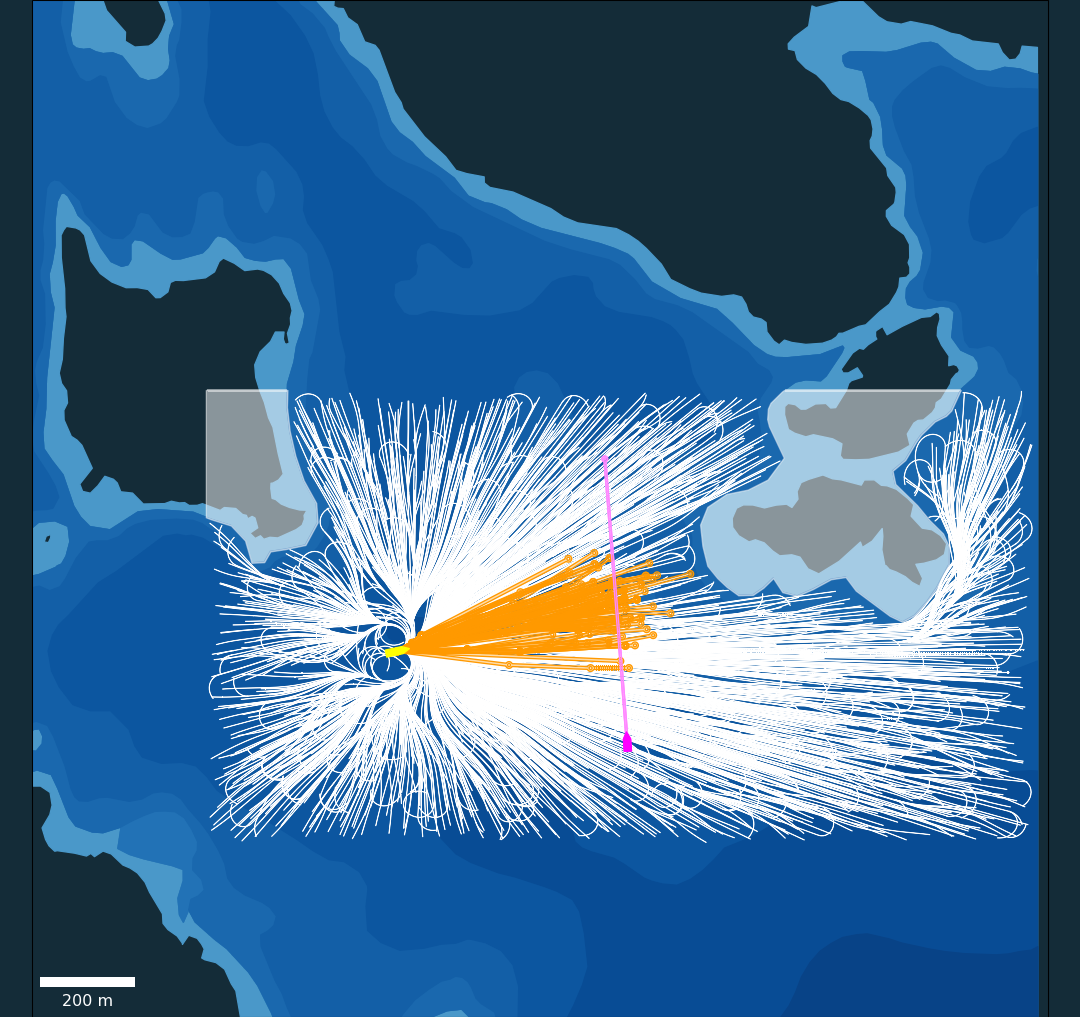}}
    \subcaptionbox{RRT*. \label{fig:rrt_star_scengen2}}
    {\includegraphics[width=0.45\columnwidth,trim={0em, 0em, 0em, 0em},clip]{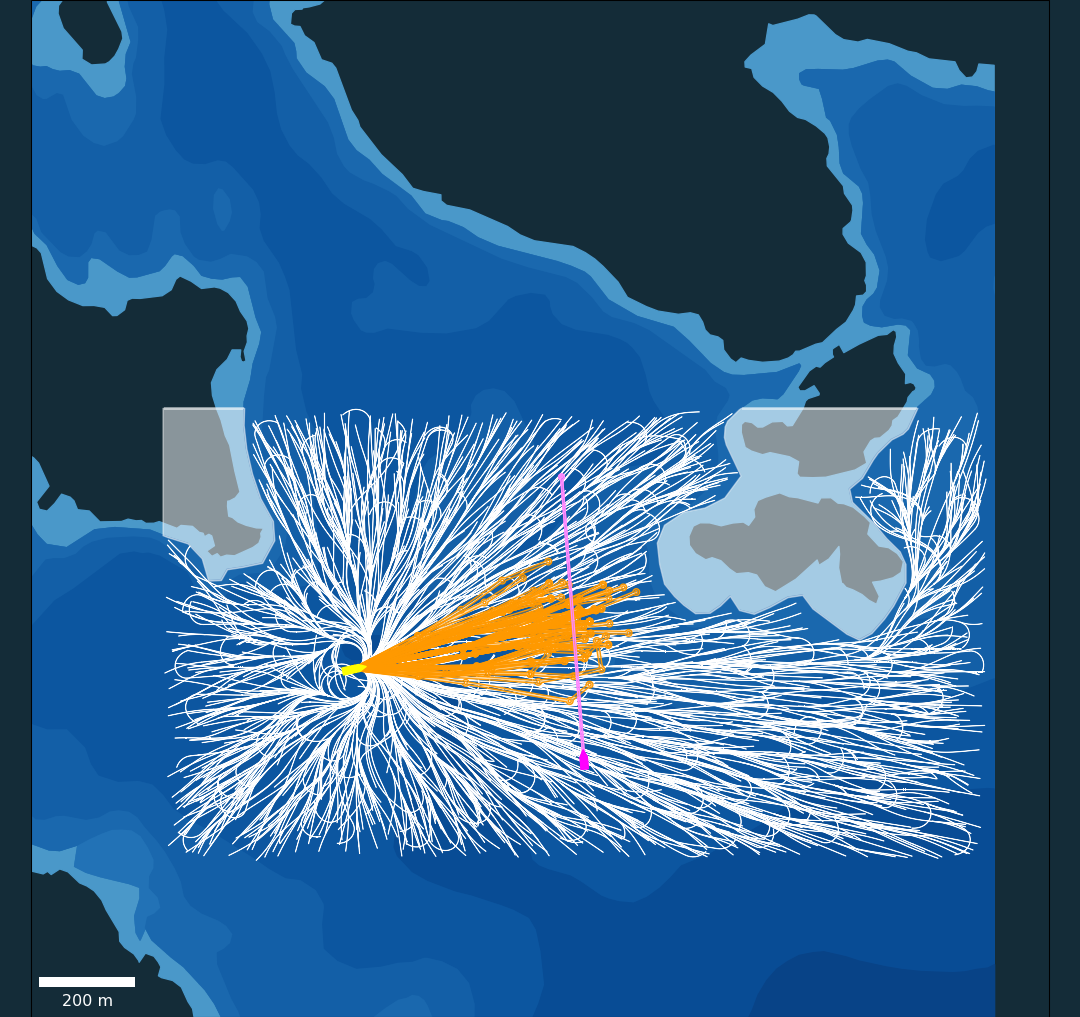}}
    \subcaptionbox{RRT. \label{fig:rrt_scengen2}}
    {\includegraphics[width=0.45\columnwidth,trim={0em, 0em, 0em, 0em},clip]{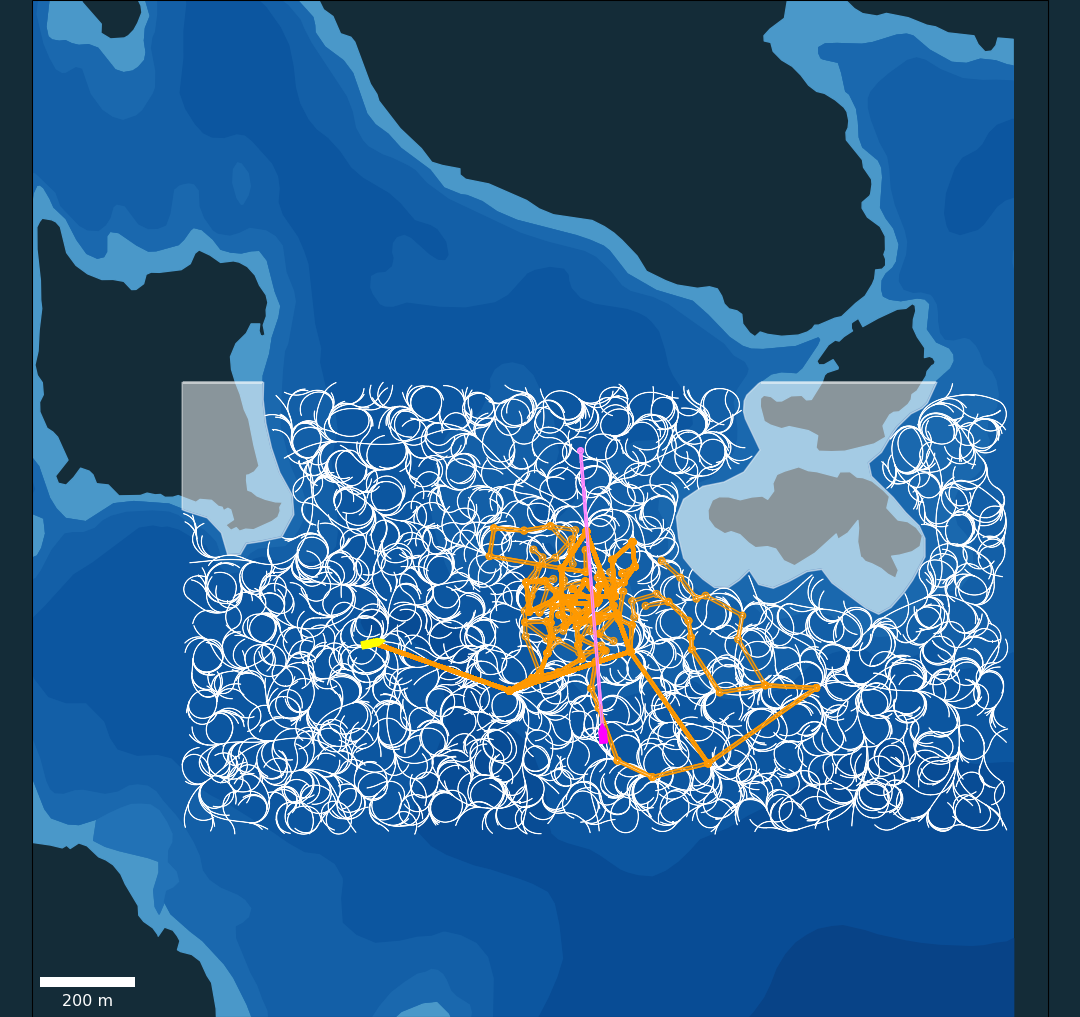}}\\
    \caption{Results from random obstacle ship behavior generation for a CAS crossing situation near B\o r\o y in Boknafjorden, Norway. The initial obstacle ship position is shown in yellow, the initial own-ship position in \textcolor{CarnationPink}{pink} and its waypoints in \textcolor{Purple}{purple}, the resulting RRT in white, sampled obstacle ship behavior waypoint sequences in \textcolor{Orange}{orange} annuluses. The safety buffered hazard boundary is shown in \textcolor{Red}{red}, with the planning bounding box shown in transparent white. }
    \label{fig:scen_gen_ex2}
\end{figure}

Fig \ref{fig:scen_gen_ex2} shows another case where we sample behaviors for a CAS crossing situation. In this situation, random position samples are drawn near the predicted closest point of approach (CPA) between the vessels, from which the nearest RRT behaviors or trajectories are fetched, i.e. 
\begin{equation}
    \begin{aligned}
        \bm{p}_{rand} &= \mathrm{Normal}(\bm{p}; \bm{p}_{cpa}, \bm{\Sigma}), 
    \end{aligned}
    \label{eq:cpa_sample}
\end{equation}
where $\bm{p}_{cpa}$ is the target ship position at CPA, assuming constant speed and course for the two vessels, calculated as in e.g. \citep{Kuwata2014}, and where $\bm{\Sigma}$ is the covariance parameter adjusting the spread of samples. This strategy allows for generating multiple target ship behaviors that will lead to a collision or near miss with the own-ship unless preventive actions are taken. Since the RRTs are flexible, any of the navigational risk factors as outlined in \citep{Bolbot2022automatic} could be considered as targets for developing either RRT sampling schemes or cost functions.

For each of the example situations, the sampled behaviors parameterized by waypoints are shown. Since the planners consider the underlying ship model dynamics, the waypoints will be feasible, in addition to being collision-free with respect to nearby static hazards. Note that we can also use the trajectory that accompanies the waypoints as well. Further note that we define a reduced-size bounding box considered by the RRT planners, in order to reduce computation time and consider only a subset of the grounding hazards present in the ENC. Once the RRTs are built, a new behavior can be sampled in less than $\SI{1}{ms}$ on the considered computing platform, which makes the approach viable for large-scale scenario production. We also note that the trees can easily be built offline, and then effectively sampled from afterwards in the relevant context.

\subsection{Smaller Planning Example}\label{subsec:case2}
In the second case, we consider a smaller planning situation with a constant reference ship speed $U_d = \SI{4.0}{m\per s}$, and allow the planners to find and refine their solution over the maximum allowable iterations and tree nodes up until a maximum time of $\SI{50}{s}$. The considered location near Kvits\o y in Rogaland, Norway has a map size of $\SI{850}{m} \times \SI{750}{m}$, and features a typical local minimum problem where one can get stuck in the dead end not far from the ship initial position. Note that as the planners sample from a CDT constructed from the safe sea area, the sampling efficiency will be higher, making it easier to avoid local minima issues.

\begin{figure*}[htp]
    \centering
    \subcaptionbox{PQ-RRT*. \label{fig:pqrrt_res1}}
    {\includegraphics[width=0.45\columnwidth,trim={0em, 0em, 0em, 0em},clip]{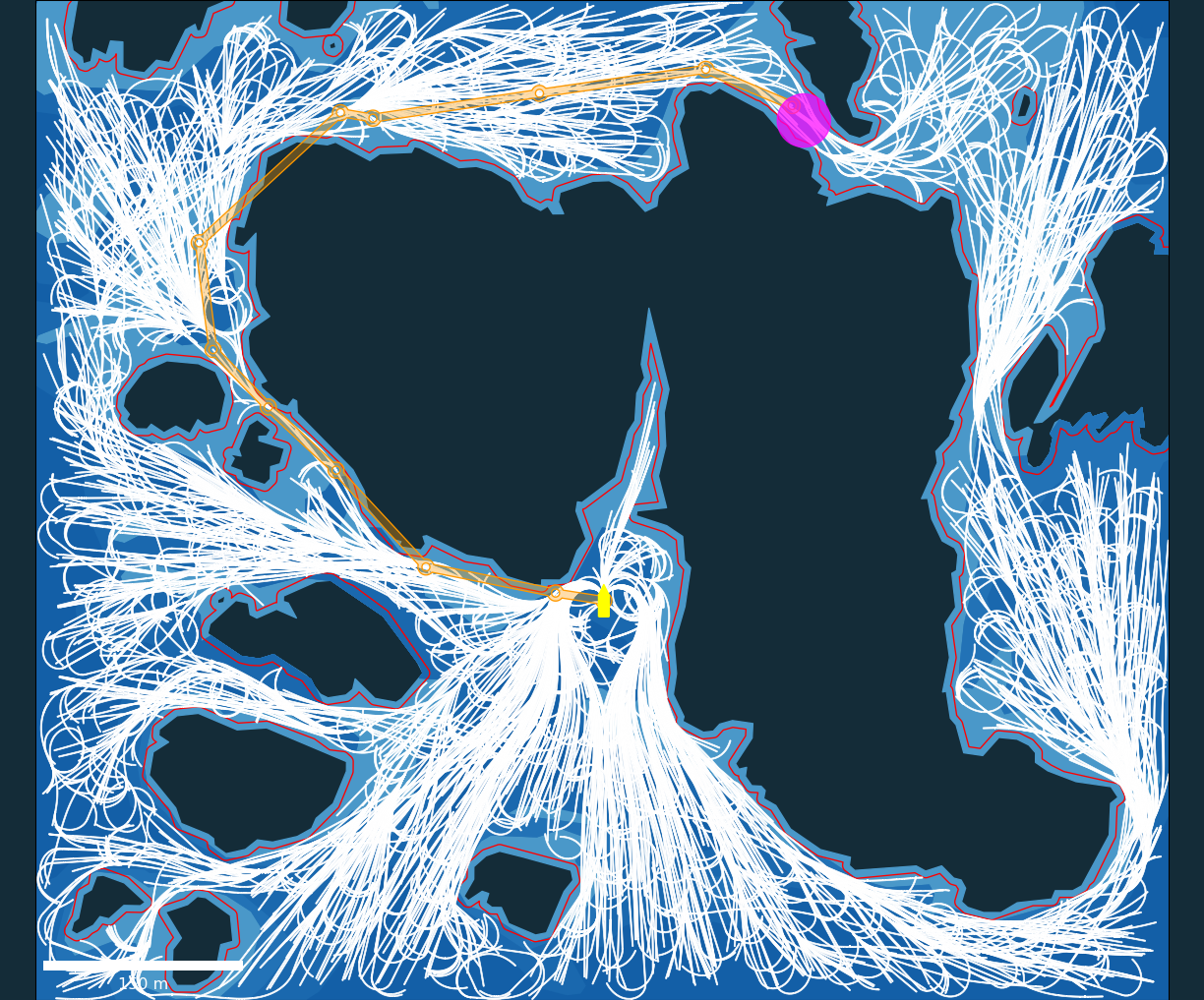}}
    \subcaptionbox{IRRT*. \label{fig:informed_rrt_res1}}
    {\includegraphics[width=0.45\columnwidth,trim={0em, 0em, 0em, 0em},clip]{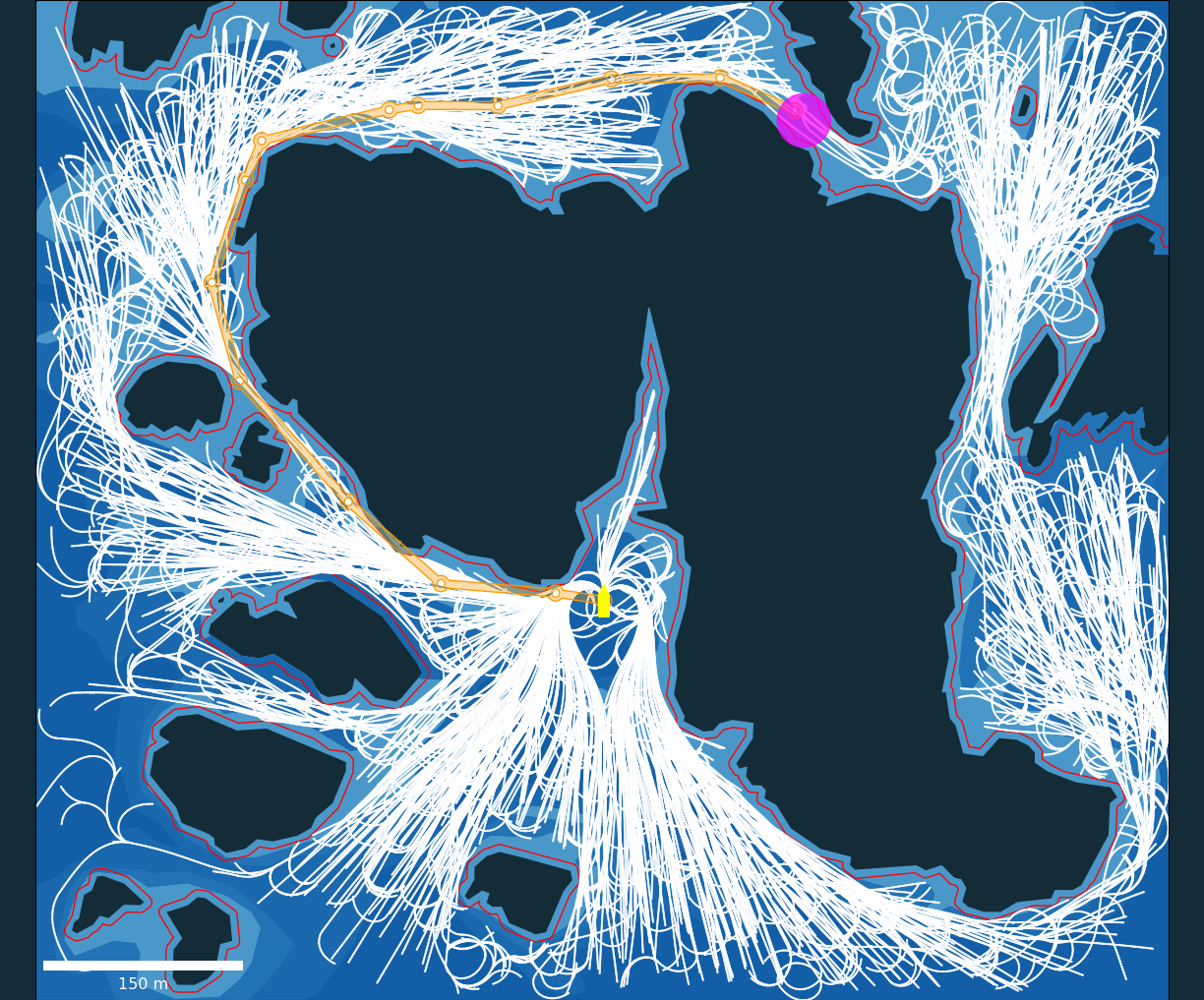}}\\
    \subcaptionbox{RRT*. \label{fig:rrt_star_res1}}
    {\includegraphics[width=0.45\columnwidth,trim={0em, 0em, 0em, 0em},clip]{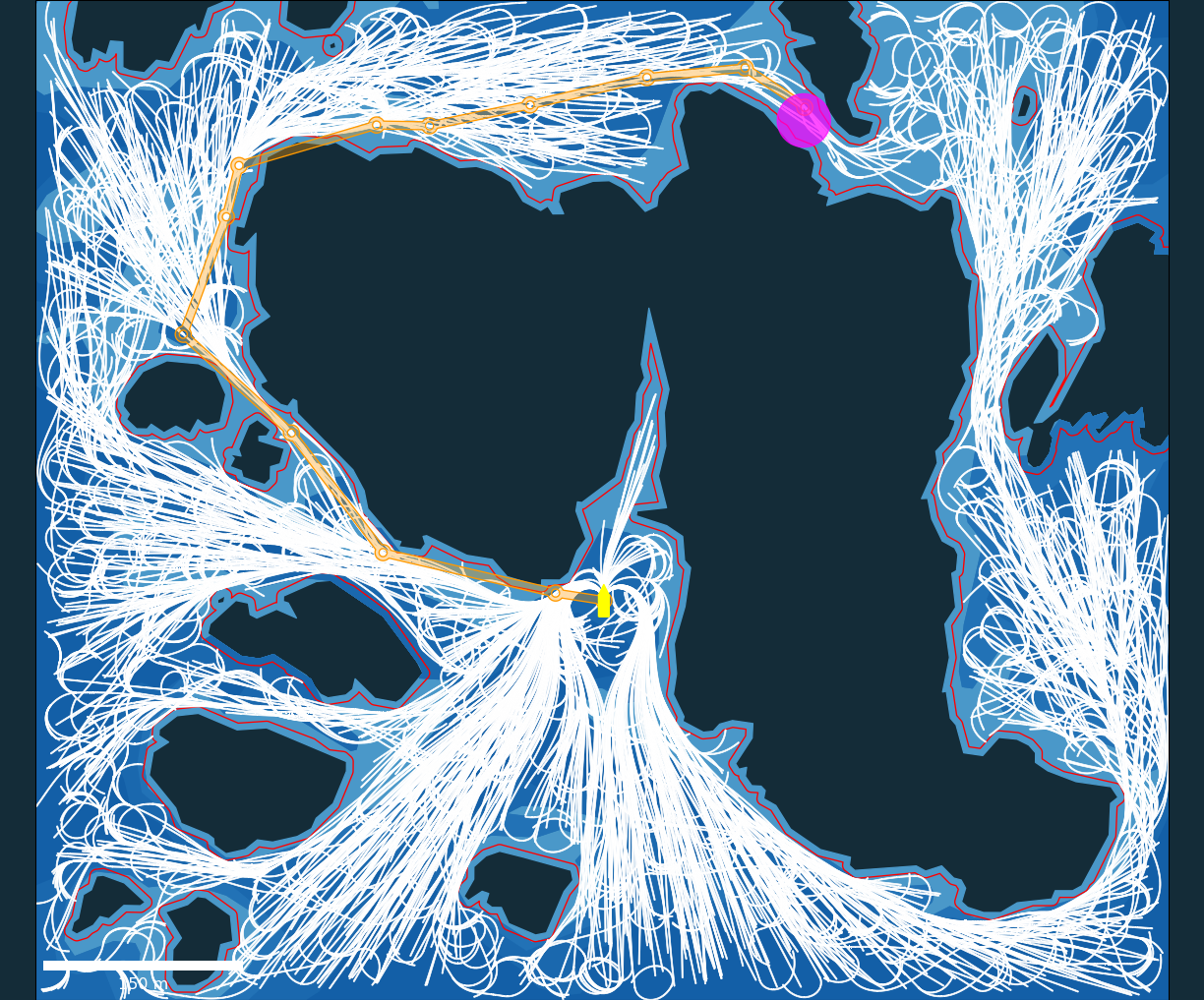}}
    \subcaptionbox{RRT. \label{fig:rrt_res1}}
    {\includegraphics[width=0.45\columnwidth,trim={0em, 0em, 0em, 0em},clip]{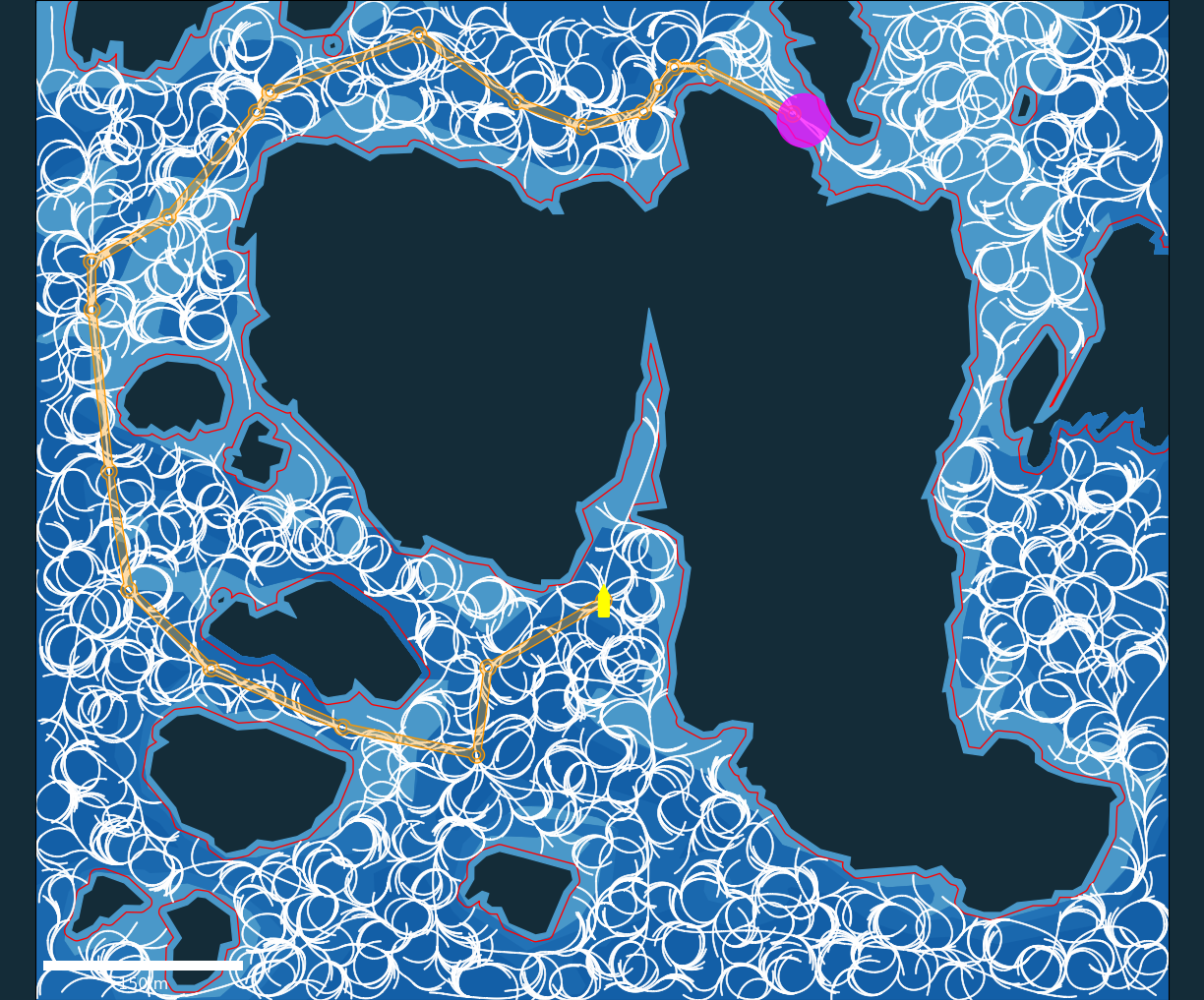}}
    \caption{Results from the first MC sample run on the smaller planning case near Kvits\o y outside Stavanger, Norway. The starting ship pose is shown in yellow, the resulting tree in white, node waypoints in \textcolor{Orange}{orange}, and the goal in \textcolor{Purple}{purple}. The safety buffered hazard boundary is shown in \textcolor{Red}{red}. }
    \label{fig:small_planning}
\end{figure*}

Results from sample runs are shown in Figure \ref{fig:small_planning}, whereas solution statistics are given in Table \ref{tab:case_2_results}. In the table, statistics for the time to find an initial solution $t_{sol,0}$ are reported, whereas run-time $t_{sol}$ and path length $d_{sol}$ statistics are reported for the final refined trajectory. The metric $\rho_{mc}$ is the success percentage of finding a solution out of all the MC runs. From Figure \ref{fig:small_planning}, it might look like there is a collision due to the orange waypoints crossing a hazard at some point. This is not the case, as the actual ship trajectories wired by the RRTs are collision-free, taking the nonholonomic properties of the ship into account. 

The optimal solution has a path length of approximately $\SI{905}{m}$. Thus we see that PQ-RRT*, IRRT*, and RRT* can converge to within $6\%$ of the optimum. PQ-RRT* attains the best results concerning path length, with a marginal difference to IRRT* and RRT*. On the other hand, the ancestor consideration in the tree rewiring and extra sampling functionality comes at the cost of higher runtimes. We also see a factor of 10 increase in the run-time between baseline RRT and RRT*, which is expected due to the more complicated wiring process. 

For IRRT* we can get marginally better results than RRT* at lower run-times. This is due to the high sample rejection rate after finding an initial solution, as samples are likely to be taken from $\mathcal{X}_{obst}$ in this map with large hazard coverage. However, accepted samples will have higher values, partially explaining the marginal improvement in solution lengths. It can here be a viable approach to iteratively prune out triangles in the safe sea CDT as the hyperellipsoid forming the sampling heuristic reduces in volume. One should thus implement efficient methods for re-computing the CDT from the new sampling domain or for pruning CDT triangles outside the domain.

\begin{table*}[htp]
	\centering
	\caption{Results for the smaller planning case over $N_{mc} = 100$ MC runs.}
	\bgroup
	\def\arraystretch{1.4}%
	\begin{tabular}{c|c|c|c|c}
		\text{Metric} & \text{PQ-RRT*} & \text{IRRT*} & \text{RRT*} & \text{RRT} \\
            \hline
		$t_{sol, 0}$	& $5.5 \pm \SI{5.0}{s}$ & $3.5 \pm \SI{3.4}{s}$ & $2.9 \pm \SI{1.8}{s}$ & $0.14 \pm \SI{0.07}{s}$ \\
            \hline
		$(t_{sol, 0}^{min}, t_{sol, 0}^{max})$	& $(\SI{0.2}{s}, \SI{29.1}{s})$ & $(\SI{0.2}{s}, \SI{15.4}{s})$ & $(\SI{0.5}{s}, \SI{11.2}{s})$ & $(\SI{0.03}{s}, \SI{0.45}{s})$ \\
		\hline
		$t_{sol}$	& $34.6 \pm \SI{2.5}{s}$ & $26.3 \pm \SI{0.7}{s}$ & $32.6 \pm \SI{0.5}{s}$ & $1.91 \pm \SI{0.03}{s}$ \\
            \hline
		$(t_{sol}^{min}, t_{sol}^{max})$	& $(\SI{28.8}{s}, \SI{38.9}{s})$ & $(\SI{24.6}{s}, \SI{28.8}{s})$ & $(\SI{31.5}{s}, \SI{34.1}{s})$ & $(\SI{1.8}{s}, \SI{2.0}{s})$ \\
            \hline 
            $d_{sol}$	& $963.1 \pm \SI{32.0}{m}$ & $965.9 \pm \SI{31.0}{m}$ & $968.3 \pm \SI{33.3}{m}$  & $1472.7 \pm \SI{163.7}{m}$ \\
            \hline 
            $(d_{sol}^{min}, d_{sol}^{max})$	& $(\SI{914.0}{m}, \SI{1081.9}{m})$ &  $(\SI{920.0}{m}, \SI{1072.0}{m})$ &  $(\SI{909.9}{m}, \SI{1070.0}{m})$ &  $(\SI{1200.0}{m}, \SI{2013.8}{m})$ \\
            \hline 
            $\rho_{mc}$	& $100 \%$ & $100 \%$ & $100 \%$ & $100 \%$ \\
	\end{tabular}
	\egroup
	\label{tab:case_2_results}
\end{table*}

The statistical significance of the results is gauged by using Welch`s $\mathit{t}$-test \citep{Ruxton2006unequal}, considering the null hypothesis $H_0$ that the mean Euclidean trajectory length from PQ-RRT* is less than that of all the other RRT planners, and the alternative hypothesis $H_1$ that the trajectory length of PQ-RRT* is equal to or larger than the other RRT planners:
\begin{equation}
\begin{split}
    H_0:  \mu_{p, \text{PQ-RRT*}} - \mu_{p, i} < 0, \qquad i \in \{\text{RRT}, \text{RRT*}, \text{IRRT*}\} \\ 
    H_1: \mu_{p, \text{PQ-RRT*}} - \mu_{p, i} \geq 0, \qquad i \in \{\text{RRT}, \text{RRT*}, \text{IRRT*}\}
\end{split}
\label{eq:hyp_test}
\end{equation}
where $\mu_{p, \text{PQ-RRT*}}$ is the mean trajectory length produced by the PQ-RRT* planner (equal to $\SI{963.1}{m}$ in the smaller planning example) and $\mu_{p, i}$ the mean trajectory length for the other RRT planners.

We start by computing Welch`s test statistic from the sampled data as
\begin{equation}
    \mathit{t}_{s, i} = \frac{\mu_{p, \text{PQ-RRT*}} - \mu_{p, i}}{\mathit{s}_i}
\end{equation}
which follows Student`s $t$-distribution under the null hypothesis $H_0$, where the pooled standard deviation $\mathit{s}_i$ is computed as 
\begin{equation}
    \mathit{s}_i = \frac{1}{\sqrt{N_{MC}}} \sqrt{\varepsilon_{p, \text{PQ-RRT*}}^2 + \varepsilon_{p, i}^2}, \qquad i \in \{\text{RRT}, \text{RRT*}, \text{IRRT*}\}
\end{equation}
To check whether the null hypothesis $H_0$ holds, we use a significance level of $\alpha=0.05$ and compute the $p$-values $P(\mathit{t}_{s, i} \geq \mathit{t}_{1-\alpha, i} | H_0)$ under the null hypothesis for the observed test statistics. This is done by utilizing the cumulative Student`s $t$-distribution function with $N_{DOF, i}$ degrees of freedom, computed through the Welch–Satterthwaite equation \citep{Ruxton2006unequal}
\begin{equation}
    N_{DOF, i} = N_{MC}^2(N_{MC} - 1)\frac{\mathit{s}_i^4}{\varepsilon_{p, \text{PQ-RRT*}}^4 + \varepsilon_{p, i}^4},
\end{equation}
which is valid for the case when the sample sizes of the two means are equal. Here, $\varepsilon_{\cdot}$ is the standard deviation of the trajectory length for the relevant planner (equal to $\SI{32.0}{m}$ in the smaller planning example for PQ-RRT*). Then, the threshold value used to determine whether we reject the null hypothesis or not is given by $\mathit{t}_{1-\alpha, i} = \mathrm{tDistributionCDF}(1 - \alpha, N_{DOF, i})$, where $\mathrm{tDistributionCDF}$ is the Student`s $t$ cumulative distribution function. The results are summarized in Table \ref{tab:case_2_p_val_results}, and show that the null hypothesis holds under the significance level of $\alpha = 0.05$ since all the $p$-values are non-significant ($> \alpha$).
\begin{table*}[htp]
	\centering
	\caption{Hypothesis testing results for the trajectory length difference of PQ-RRT* versus the other planners in the smaller planning case.}
	\bgroup
	\def\arraystretch{1.4}%
	\begin{tabular}{c|c|c|c}
		\text{} &  $i=\text{IRRT*}$ & $i=\text{RRT*}$ & $i=\text{RRT}$ \\
            \hline
		$\mathit{t}_{s, i}$	& $-0.6285$ & $-1.1260$ & $-30.5519$ \\
            \hline
		$\mathit{s}_{i}$	& $4.4553$ & $4.6183$ & $16.6798$ \\
		\hline
		$N_{DOF,i}$	& $ 197$ & $197$ & $106$ \\
            \hline
		$\mathit{t}_{1-\alpha, i}$	& $1.6526$ & $1.6526$ & $1.6526$ \\
            \hline 
            $p$-value	& $0.7348$ & $0.8692$ & $1.0$ \\
	\end{tabular}
	\egroup
	\label{tab:case_2_p_val_results}
\end{table*}

\subsection{Larger Planning Example}
The larger planning example considers the region shown in Fig. \ref{fig:cdt_ex} of size $\SI{5.0}{km} \times \SI{5.0}{km}$ with multiple islands and smaller grounding hazards, which increases the problem size substantially. Again, the planners are allowed to find and refine a solution over the maximum allowable iterations and nodes, up until a maximum time of $\SI{300}{s}$. A constant reference speed $U_d = \SI{5.0}{m\per s}$ is utilized. In this case, due to the large map size, we utilize planner parameters as in Table \ref{tab:parameters_case_3}.
\begin{table}[!htp]
	\centering
	\caption{Algorithm parameters for the larger planning case.}
	\bgroup
	\def\arraystretch{1.4}%
	\begin{tabular}{c|c|c|c|c}
		\text{Parameter} & \text{PQ-RRT*} & \text{Informed-RRT*} & \text{RRT*} & \text{RRT} \\
		\hline
		$N_{node}^{max}$	& $10000$ & $10000$ & $10000$ & $10000$ \\
            \hline 
            $N_{iter}^{max}$	& $25000$ & $25000$ & $25000$ & $25000$ \\
            \hline 
            $\Delta_{goal}$	& $500$ & $500$ & $500$ & $500$ \\
            \hline 
            $d_{node}^{min}$	& $\SI{15.0}{m}$ & $\SI{15.0}{m}$ & $\SI{15.0}{m}$ & - \\
            \hline 
            $\gamma$ & $\SI{3500.0}{m}$ & $\SI{3500.0}{m}$ & $\SI{3500.0}{m}$ & - \\
            \hline 
            $T_{min} $ & $\SI{1.0}{s}$ & $\SI{1.0}{s}$ & $\SI{1.0}{s}$ & $\SI{1.0}{s}$ \\
		\hline
            $T_{max} $ & $\SI{30.0}{s}$ & $\SI{30.0}{s}$ & $\SI{30.0}{s}$ & $\SI{30.0}{s}$ \\
            \hline
            $R_{a}$	& $\SI{10.0}{m}$ & $\SI{10.0}{m}$ & $\SI{10.0}{m}$ & $\SI{10.0}{m}$ \\
            \hline 
            $\delta_{sim} $ & $\SI{1.0}{s}$ & $\SI{1.0}{s}$ & $\SI{1.0}{s}$ & $\SI{1.0}{s}$ \\
            \hline
            $d_{margin}$ & $\SI{0.5}{m}$ & - & - & - \\ 
            \hline
            $N_{ancestry}$ & $1$ & - & - & - \\ 
            \hline
            $N_{sa}^{max}$ & $50$ & - & - & - \\
            \hline
            $\lambda_{sample}$ & $\SI{8.0}{m}$ & - & - & - \\
	\end{tabular}
	\egroup
	\label{tab:parameters_case_3}
\end{table}

\begin{figure*}[htp]
    \centering
    \subcaptionbox{PQ-RRT*. \label{fig:pqrrt_res}}
    {\includegraphics[width=0.45\columnwidth,trim={0em, 0em, 0em, 0em},clip]{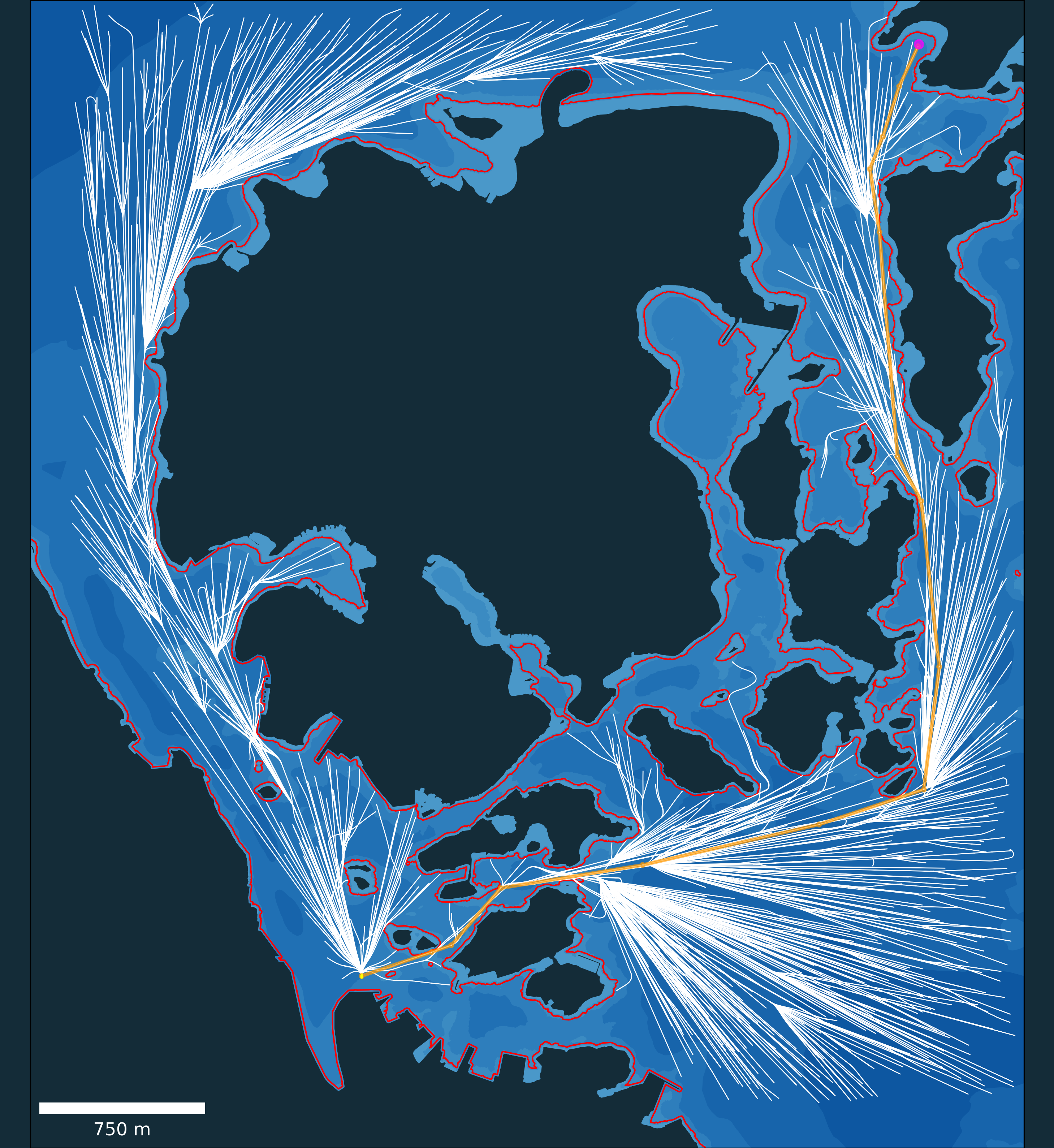}}
    \subcaptionbox{IRRT*. \label{fig:informed_rrt_res}}
    {\includegraphics[width=0.45\columnwidth,trim={0em, 0em, 0em, 0em},clip]{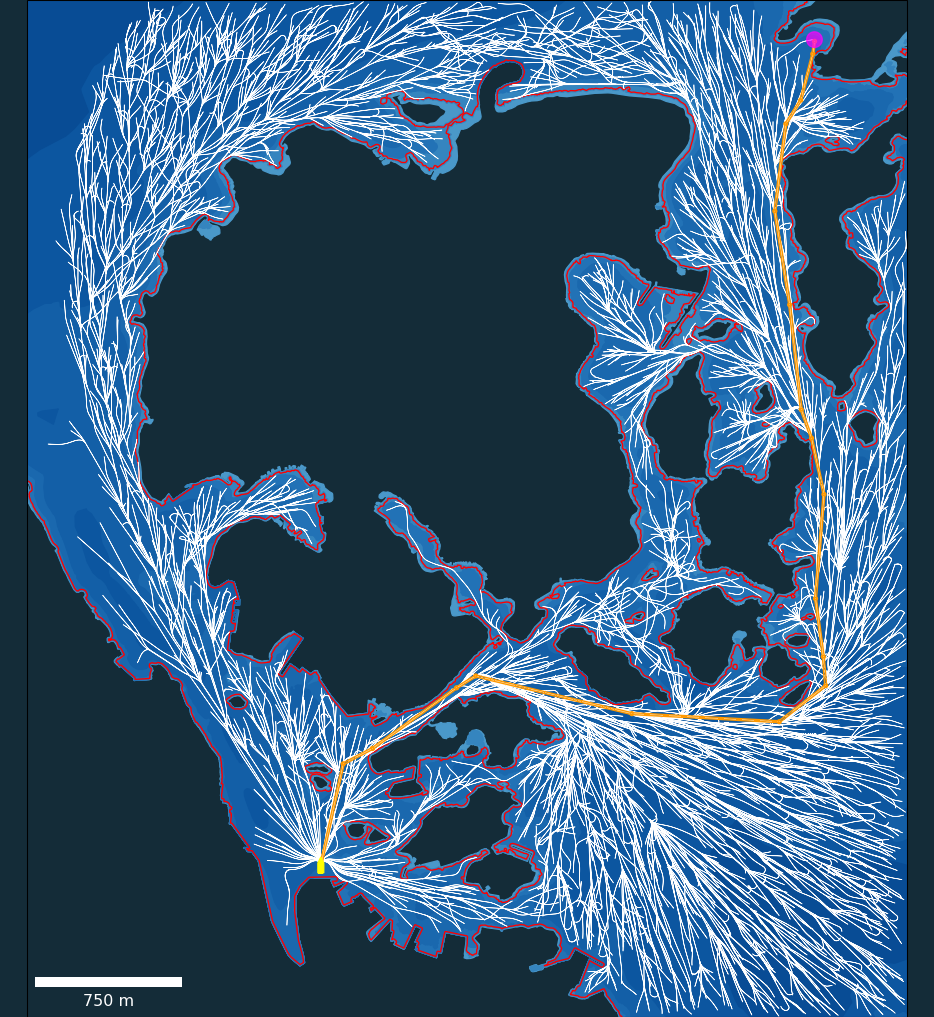}}\\
    \subcaptionbox{RRT*. \label{fig:rrt_star_res}}
    {\includegraphics[width=0.45\columnwidth,trim={0em, 0em, 0em, 0em},clip]{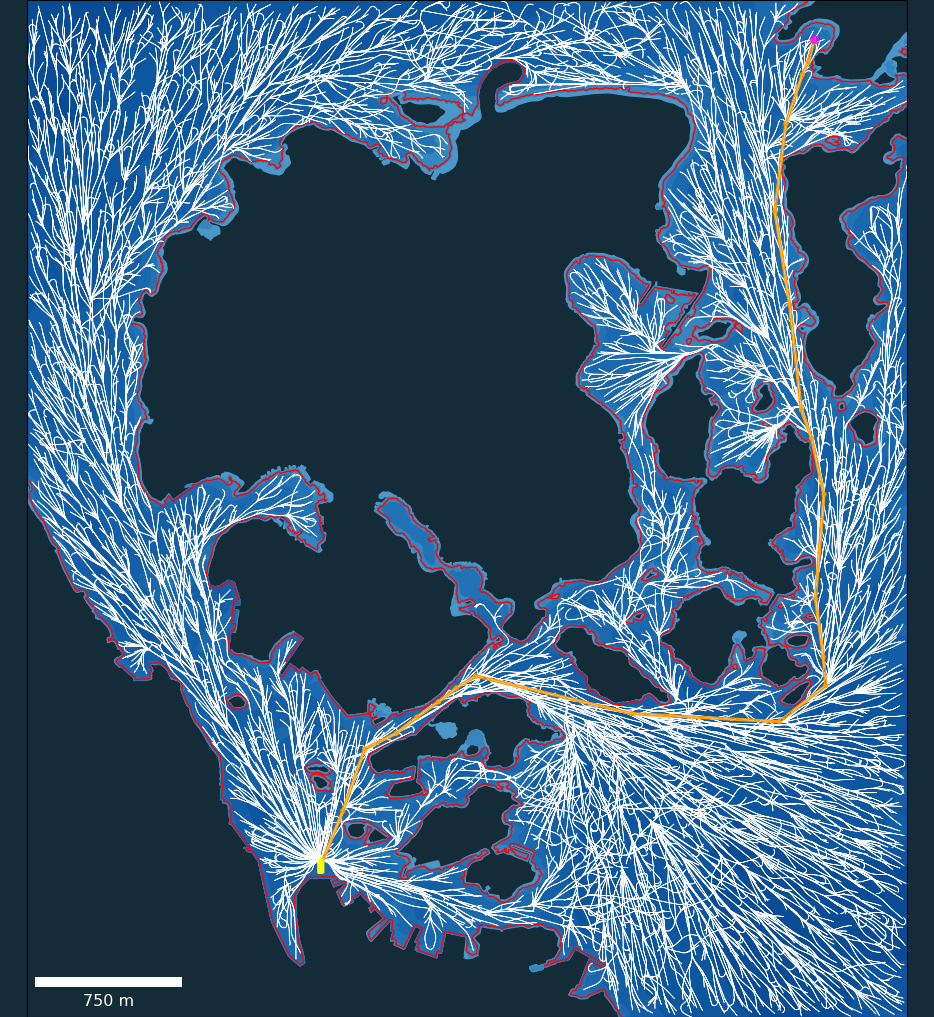}}
    \subcaptionbox{RRT. \label{fig:rrt_res}}
    {\includegraphics[width=0.45\columnwidth,trim={0em, 0em, 0em, 0em},clip]{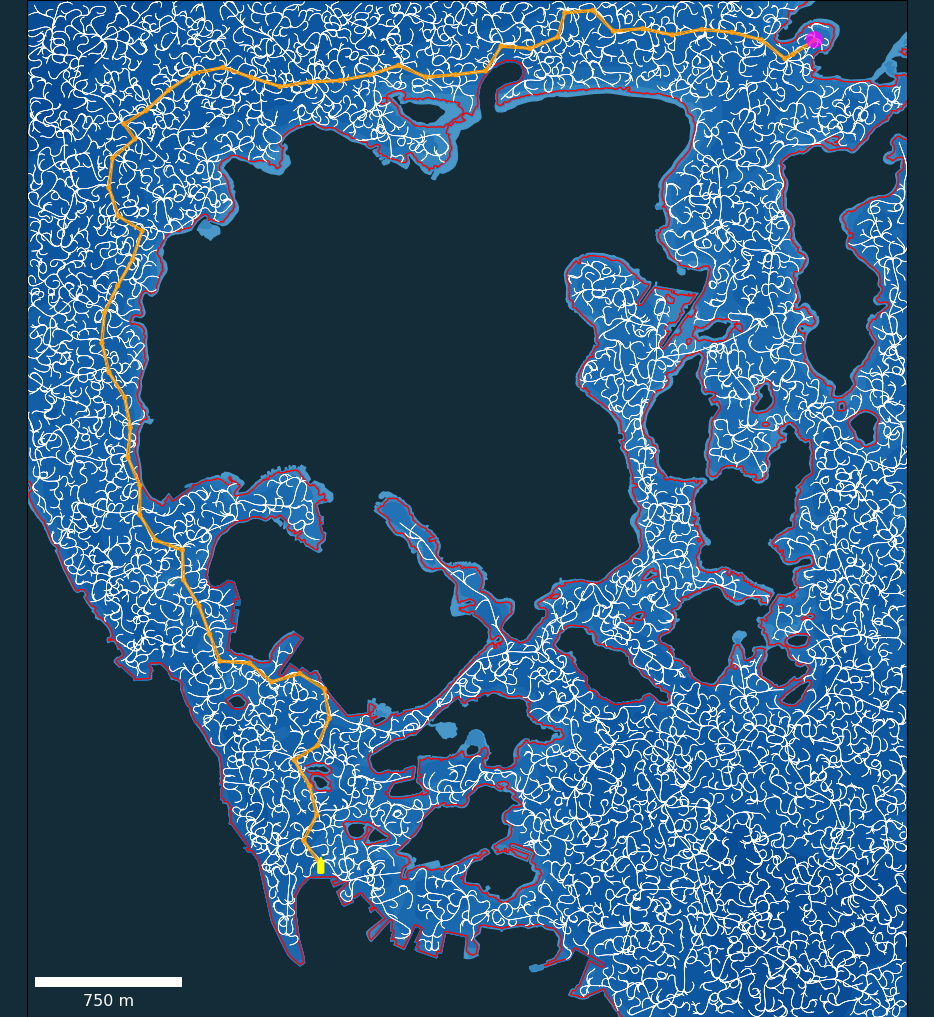}}
    \caption{Results from the first MC sample run on the larger planning case near V\aa gen in Stavanger, Norway. The starting ship pose is shown in yellow, the resulting tree in white, node waypoints in \textcolor{Orange}{orange}, and the goal in \textcolor{Purple}{purple}. The safety buffered hazard boundary is shown in \textcolor{Red}{red}. }
    \label{fig:large_planning_1}
\end{figure*}

Visual results when applying the RRT-variants on the planning area in Fig. \ref{fig:cdt_ex} are shown for sample runs in Fig. \ref{fig:large_planning_1}. Table \ref{tab:case_3_results} shows performance metrics for the algorithms over the MC runs. Again, we see a marginally better result for the PQ-RRT* than for IRRT* and RRT*. In this planning example, IRRT* achieves worse results than RRT* due to the significantly higher sample rejection rate again caused by a large obstacle congestion ratio, amplified by the problem scale.

The optimal solution is approximately $\SI{5.2}{km}$, and thus we see that the optimal algorithm variants only converge to within approximately $30\%$ of the optimum. The convergence issue is attributed to the map size, the optimal solution passing through narrow passages, and the complexity of considering ship dynamics and kinodynamical constraints in the tree wiring. A standard deviation of over $\SI{500}{m}$ is found for the path length solutions of all variants, which is significantly high.

\begin{table*}[htp]
	\centering
	\caption{Results for the larger planning case over $N_{mc} = 100$ MC runs.}
	\bgroup
	\def\arraystretch{1.4}%
	\begin{tabular}{c|c|c|c|c}
		\text{Metric} & \text{PQ-RRT*} & \text{IRRT*} & \text{RRT*} & \text{RRT} \\
            \hline
		$t_{sol, 0}$	& $62.7 \pm \SI{22.3}{s}$ & $60.7 \pm \SI{31.0}{s}$ & $60.7 \pm \SI{31.0}{s}$ & $3.8 \pm \SI{1.4}{s}$ \\
            \hline
		$(t_{sol, 0}^{min}, t_{sol, 0}^{max})$	& $(\SI{33.7}{s}, \SI{162.5}{s})$ & $(\SI{19.0}{s}, \SI{205.0}{s})$ & $(\SI{19.0}{s}, \SI{205.0}{s})$ & $(\SI{1.6}{s}, \SI{9.6}{s})$ \\
		\hline
		$t_{sol}$	& $300.0 \pm \SI{0.0}{s}$ & $300.0 \pm \SI{0.0}{s}$ & $300.0 \pm \SI{0.0}{s}$ & $24.9 \pm \SI{2.1}{s}$ \\
            \hline
		$(t_{sol}^{min}, t_{sol}^{max})$	& $(\SI{300.0}{s}, \SI{300.0}{s})$ & $(\SI{300.0}{s}, \SI{300.0}{s})$ & $(\SI{300.0}{s}, \SI{300.0}{s})$ & $(\SI{21.6}{s}, \SI{27.5}{s})$ \\
            \hline 
            $d_{sol}$	& $6567.3 \pm \SI{407.7}{m}$ & $6836.6 \pm \SI{476.6}{m}$ & $6744.6 \pm \SI{487.4}{m}$  & $8805.3 \pm \SI{620.8}{m}$ \\
            \hline 
            $(d_{sol}^{min}, d_{sol}^{max})$	& $(\SI{5239.9}{m}, \SI{7359.9 }{m})$ &  $(\SI{5299.9}{m}, \SI{7534.9}{m})$ &  $(\SI{5269.9}{m}, \SI{7440.0}{m})$ &  $(\SI{6484.0}{m}, \SI{10178.5}{m})$ \\
            \hline 
            $\rho_{mc}$	& $100 \%$ & $100 \%$ & $100 \%$ & $100 \%$ \\
	\end{tabular}
	\egroup
	\label{tab:case_3_results}
\end{table*}
Again we check the statistical significance of PQ-RRT* giving more distance-optimal trajectories than the other variants, by using a one-sided Welch`s $t$-test as in the previous section, with the hypotheses in \eqref{eq:hyp_test}. The results are summarized in Table \ref{tab:case_3_p_val_results}, and again it is shown that the null hypothesis holds under the significance level of $\alpha = 0.05$ since all the $p$-values are non-significant.
\begin{table*}[htp]
	\centering
	\caption{Hypothesis testing results for the trajectory length difference of PQ-RRT* versus the other planners in the larger planning case.}
	\bgroup
	\def\arraystretch{1.4}%
	\begin{tabular}{c|c|c|c}
		\text{} &  $i=\text{IRRT*}$ & $i=\text{RRT*}$ & $i=\text{RRT}$ \\
            \hline
		$\mathit{t}_{s, i}$	& $-4.2938$ & $-2.7902$ & $-30.1331$ \\
            \hline
		$\mathit{s}_{i}$	& $62.7190$ & $63.5435$ & $74.2706$ \\
		\hline
		$N_{DOF,i}$	& $ 193$ & $192$ & $171$ \\
            \hline
		$\mathit{t}_{1-\alpha, i}$	& $1.6528$ & $1.6528$ & $1.6538$ \\
            \hline 
            $p$-value	& $0.9999$ & $0.9971$ & $1.0$ \\
	\end{tabular}
	\egroup
	\label{tab:case_3_p_val_results}
\end{table*}

\subsection{Discussion}
Gauging the results on planning, we see that the RRT-based planners are viable for use in problems of adequate size, i.e. less than $\SI{1}{km} \times \SI{1}{km}$ in map size. In these cases, the planners can find and optimize the best solution in adequate time. However, for larger maps, the planners struggle. This is attributed to the increased number of samples and iterations required in order to find and refine a solution. Although the planners find initial solutions fast, they have a hard time optimizing the solutions when considering larger map sizes and complex environments. The run-time also increases substantially for larger problem sizes, due to the higher computational cost of re-wiring the tree and propagating new node costs to the leaves. Thus, in such large cases, it can be an option to utilize RRT-based planners without dynamics consideration, where the tree wiring only considers position sampling and the connection of these through collision-free straight-line segments. Alternatively, one can use path complete algorithms such as A*, to find the solution fast up to a suitable grid resolution, without having randomness in the result. Note that it will then be necessary to post-process the solution to generate a feasible trajectory for the ship to track, unless a motion primitives-based planner is used. 

We note that IRRT*, although providing linear algorithm convergence properties for obstacle-free environments, struggles with both planning cases and especially the larger one. This is due to the large volume occupied by obstacles in the configuration space, leading to a significant rejection rate in the informed heuristical sampling. Thus, for IRRT* to be of practical usage, it requires an improvement. This can be an update and creation of a new CDT for the safe sea domain inside the informed sampling domain, each time a new solution is found.

For the PQ-RRT* with nonholonomic steering, we see a large increase in computational effort due to the ancestry consideration. This was viable for the smaller planning case, but proved to be more limiting in the larger case unless the nonholonomic steering functionality is disabled or higher simulation time steps are used. Also, the algorithm run-time and performance are highly dependent on the sample adjustment procedure, itself being dependent on the three parameters $N_{sa}^{max}$, $\lambda_{sa}$, and $d_{margin}$. In total, this leads to a lower tree node number and can in the worst cases prevent the algorithm from finding solutions. Choosing $\lambda_{sa}$ too small yields negligible gain from the APF-based adjustments, and requires a higher iteration number $N_{sa}^{max}$, which again gives higher algorithm run-time due to an increased number of obstacle distance calculations being required. Conversely, a large $\lambda_{sa}$ can lead to rapid convergence towards local minima. We found that a selection of $\lambda_{sa}$ in the order of $0.1\%$ of the map width, and an adjustment number of around $N_{sa}^{max} \approx 50$ gave a reasonable trade-off between run-time and solution quality for the larger planning case in this work. For the clearance parameter $d_{margin}$ it was found necessary to select a smaller value of around $\SI{0.5}{m}$, as a high value can prevent the planner from wiring into more confined areas. Because of these factors, we found the tuning of PQ-RRT* to be significantly more challenging than the other variants. Judging from the marginal improvements found in this work with consideration of nonholonomic steering, we argue that it was easier to employ IRRT* or RRT*, which required less tuning effort and achieved comparable performance. In general, we note that the parameter selection is dependent on the vehicle steering system. Another tuning challenge is the selection of $\gamma$, also mentioned in \cite{Noreen2016comparison}, which has a large influence on performance. As the ball volume reduces with the tree size, too small values of $\gamma$ can lead to negligible search radius and can in the worst case effectively reduce optimal RRT variants to the baseline RRT, where no nearest neighbors other than the closest one are considered in the wiring. A solution to consider is to bound the search radius from below, or for simplicity consider a fixed search radius.

We see that the common challenge of RRTs related to a slow convergence rate towards the optimum, is increasing when applying the algorithms to large and complex environments. This is partially due to the sampling inefficiency and node rejection rate issue found in most variants, and for which a significant amount of solutions have been proposed with varying levels of success \citep{Veras2019systematic}. Also, for real-time systems and applications where time is a resource, the incremental tree wiring and re-wiring induce a computational cost that must be weighted against performance and optimality. Thus, we deem computationally constrained RRT-based planners without sophisticated sampling strategies and without coupling to graph-search-based methods such as A*, to be more suitable for smaller problems, or problems where obstacles occupy a smaller portion of the configuration space, or where non-optimal solutions are acceptable. For planning in higher dimensional space, it is also necessary to consider other metrics than the Euclidean one \citep{Noreen2016comparison}.

\section{Conclusion}\label{sec:conclusion}
In this article, multiple algorithms for ship trajectory planning based on RRT have been developed and compared with respect to trajectory length and computational time. The comparison focuses on varying degrees of difficulty in a complex environment containing many non-convex grounding hazards, as opposed to the often simple environments used for testing in previous work. 

Practical aspects to consider when employing such algorithms in the maritime domain are also outlined and discussed, to the benefit of researchers and practitioners in the field. It is also shown through an example case that RRT variants can be beneficial in the context of automatic test scenario generation, where target ship trajectories can be sampled efficiently directly from the nodes of a built RRT. The tree can alternatively be used to sample intention scenarios for use in intelligent CAS. 

From Monte Carlo simulations on selected cases, we see that PQ-RRT* attains more distance optimal trajectories, also verified through pair-wise hypothesis testing with Welch`s $t$-test when using a significance level $\alpha=0.05$. Here, IRRT* and RRT* follow close behind. This distance optimality comes naturally at the cost of increased run-time due to nearest neighbor searches and parent consideration in both tree wiring and rewiring. IRRT* struggles with cases where obstacles cover a large part of the configuration space. In larger planning cases, more efficient sampling procedures are needed for optimal RRT* variants to be viable due to the significant space of configurations that must be covered, causing a similar curse of dimensionality issue. This causes inefficiencies in the trade-off between computational effort, available memory, and performance, as the optimal variants will then spend the majority of their effort wiring and re-wiring their tree.

From the results and through tuning of the algorithm, it was found that the PQ-RRT* involves much higher complexity in tuning than the other variants, because of the sample adjustment procedure and ancestor consideration. On the other hand, the IRRT* algorithm here attains a good balance between simpler tuning and obtainable performance. It's informed sampling heuristic should, however, be improved to reduce its sample rejection rate. In the maritime domain, this can be achieved by an iterative pruning or update of a safe sea triangulation used to sample new collision-free configurations.

\printcredits
\bibliographystyle{cas-model2-names}
\bibliography{main}
\bio{}
\endbio

\end{document}